# The Construction of Reality in an AI: A Review

**Jeffrey W. Johnston**
jeffj4a@earthlink.net

**Abstract**

AI constructivism as inspired by Jean Piaget, described and surveyed by Frank Guerin, and representatively implemented by Gary Drescher seeks to create algorithms and knowledge structures that enable agents to acquire, maintain, and apply a deep understanding of the environment through sensorimotor interactions. This paper aims to increase awareness of constructivist AI implementations to encourage greater progress toward enabling lifelong learning by machines. It builds on Guerin's 2008 "Learning Like a Baby: A Survey of AI approaches." After briefly recapitulating that survey, it summarizes subsequent progress by the Guerin referents, numerous works not covered by Guerin (or found in other surveys), and relevant efforts in related areas. The focus is on knowledge representations and learning algorithms that have been used in practice viewed through lenses of Piaget's schemas, adaptation processes, and staged development. The paper concludes with a preview of a simple framework for constructive AI being developed by the author that parses concepts from sensory input and stores them in a semantic memory network linked to episodic data. Extensive references are provided.

**Keywords**: AI constructivism, developmental robotics, lifelong learning, artificial general intelligence, infant bootstrap, general-purpose concept learning

## 1 Introduction

Constructivism in artificial intelligence (AI), inspired by the developmental psychology of Jean Piaget (1936/1954, 1952) and successors like Cohen LB (1977), Karmiloff-Smith (1992), and Mandler (1992, 2004, 2012), has been described and implemented by Cunningham (1972), Drescher (1989/1991), Shen (1989), Cohen PR et al. (1997), Kuipers (2000), Barto et al. (2004), Chaput (2004), Guerin (2008a, 2008b), and others. McLeod (2020) describes a key insight of Piaget (that is core to constructivist AI) as "Children are born with a very basic mental structure (genetically inherited and evolved) on which all subsequent learning and knowledge are based."

AI constructivists reject nativist positions that suggest innate, specialized, evolutionally programmed mechanisms underlie most mental activity. Rather, they presume agents have limited innate knowledge and general-purpose algorithms enable continuous learning in situated environments.[1] Many AI constructivists, such as Drescher (1989/1991), are aligned with developers like Albus (2008), Tenenbaum et al. (2011), Lake et al. (2016), Dupoux (2016), and DiCarlo (2018) in further characterizing the constructivist goal as "reverse engineering the human mind."

Alan Turing (1950) anticipated AI constructivism writing, "Instead of trying to produce a programme to simulate the adult mind, why not rather try to produce one which simulates the child's? If this were then subjected to an appropriate course of education one would obtain the adult brain." This approach is seen as a path to *artificial general intelligence*[2] by developers like Adams et al. (2012) who join Nilsson (2005) in advocating for "the development of general-purpose, educable systems that can learn and be taught to perform any of the thousands of jobs that humans can perform … beginning with a system that has minimal, although extensive, built-in capabilities." AI constructivism offers a promising approach for aligning agents with human sensibilities, allowing them to acquire common sense, avoiding brittleness from trying to program them for all contingencies, and enabling lifelong learning.

[1] For more discussion about nativism versus constructivism see (Karmiloff-Smith 1992; Elman et al. 1996; Johnson SP 2003; Cangelosi and Schlesinger 2015; Hutson 2018).
[2] "The intelligence of a machine that could successfully perform any intellectual task that a human being can" (Wikipedia, Artificial general intelligence).

Biology supports the constructivist approach per evidence of brain plasticity (Bach-y-Rita and Kercel 2003), neural constructivism (Quartz and Sejnowski 1997), Mountcastle's hypothesis that a single learning algorithm operates in the primate neocortex (Hawkins 2004; Dean et al. 2012), postnatal neurogenesis (Kuhn et al. 2018), and the emergence of semantic hubs in human brains (Garagnani and Pulvermuller 2016). Machine learning research has shown that nodes in artificial neural networks correspond to features and concepts induced from input streams (Carter et al. 2019; Goh et al. 2021).

This paper considers AI constructivism synonymous with *developmental robotics* (Cangelosi and Schlesinger 2015)[3], *autonomous mental development* (Weng et al. 2001; Zhang 2009)[4], *cognitive and developmental systems* (Jin 2016)[5], *epigenetic robotics* (Lungarella et al. 2003)[6], *autonomous learning* (Shen 1989, 1994), *cognitive (developmental) robotics* (Clark A and Grush 1999; Asada et al. 2009), and the *computational approach to constructivism* (Riegler 2005). It favors the term "AI constructivism" in recognition of Piaget and Drescher's pioneering work and considers "constructivism" a more evocative and precise descriptor.

Key elements of Piaget's theory of cognitive development often leveraged by AI constructivists include *schemas* (a.k.a. *schemata*) as the building blocks of skills and knowledge, *adaptation processes* of *assimilation* and *accommodation* as key learning processes[7], and cognitive development occurring in *stages*. According to McLeod (2020), Piaget considered a schema "a cohesive, repeatable action sequence possessing component actions that are tightly interconnected and governed by a core meaning." Assimilation refers to the incorporation of experiential information into existing schemas. Accommodation refers to modification of existing schemas (or creation of new ones) when experiences are incompatible with existing schemas. Piaget defined stages of development as *sensorimotor* (ages 0-2), *pre-operational* (ages 2-7), *concrete operational* (ages 7-11), and *formal operational* (ages 11+). In the sensorimotor stage, schemas are constructed from reflexes (i.e., innate schemas) that develop into increasingly purposeful behaviors via *circular reactions*.[8] Cognitive abilities developed during this stage include understanding object permanence, recognizing the self, applying behaviors observed in others to one's own behaviors (i.e., deferred imitation), and using proxies (e.g., toys) to simulate real-world situations (i.e., representational play). The pre-operational stage is characterized by development of symbolic thought. The concrete operational stage is marked by development of logical thinking. In the formal operational stage, humans develop the ability to think abstractly and engage in scientific reasoning (McLeod 2020; Wikipedia, Piaget's theory of cognitive development).

---

[3] "Developmental robotics is the interdisciplinary approach to the autonomous design of behavioral and cognitive capabilities in artificial agents (robots) that takes direct inspiration from the developmental principles and mechanisms observed in the natural cognitive systems of children (C&S 2015, p. 4)."

[4] This terminology was adopted for the *IEEE Transactions on Autonomous Mental Development* journal in 2009.

[5] *IEEE Transactions on Autonomous Mental Development* was renamed *Transactions on Cognitive and Developmental Systems* in 2016.

[6] Lungarella et al. distinguish epigenetic robotics from developmental robotics as follows: "The former focuses primarily on cognitive and social development, as well as on sensorimotor environmental interaction, the latter encompasses a broader spectrum of issues, by also investigating the acquisition of motor skills and the role played by morphological development."

[7] Piaget's later work (e.g. (Piaget 1977, 1985)) also stresses *equilibration*—a process of continuous adaptation—that balances assimilation-accomodation constructions and genetic and social influences. This paper presumes equilibration is subsumed by assimilation and accommodation. That is, when *disequilibrium* ("cognitive dissonance") occurs an existing concept must be expanded by assimilation (e.g., adding new features to a class) or new schemas created by accommodation to relieve dissonance and restore equilibrium.

[8] "Circular reaction" was used by Piaget (1952, p. 49) by way of Baldwin J (1895, p. 466) to refer to repetitive actions that become increasingly useful and purposeful as an agent develops. Merriam-Webster defines circular reaction as "a chain reflex in which the final response acts as stimulus for the initial response."

AI constructivists have adopted Piagetian elements to varying degrees. Schemas generally correspond to knowledge, skill, or concept representations. Adaptation processes correspond to *learning algorithms* that create and maintain schemas. Staged development is a key influence on the constitution of learning-algorithms and agent architectures. At its core, staging is the principle that prior knowledge and skills provide the basis for subsequent knowledge and skills. Early AI schemas include Minsky's (1975) frames, Schank and Abelson's (1975) scripts, and Rumelhart's schemas (Rumelhart and Ortony 1977). Schemas have been implemented as data structures; small modular programs; software functions, sub-routines, and objects; self-organizing maps; rules; production systems; finite state machines; semantic nets; artificial neural networks; physics simulators; mini Turing machines; combinations of these; and other constructs.

This paper is structured around a recapitulation of Guerin's (2008a) survey of constructivist systems developed by AI practitioners. Section 2 provides a brief chronological summary of the developments covered therein with a focus on data structures, algorithms, and staging. Section 3 examines progress by the Guerin referents since 2008. Section 4 presents many notable works not covered in Guerin, including some that are lesser known, out of mainstream AI, and subsequent to 2008. Section 5 lists related areas of AI research. Section 6 provides an overview of a constructivist AI approach being developed by the author that parses concepts from sensory input and stores them in a semantic memory network linked to episodic data. The approach aims to provide "mechanisms to bridge the gap between the sensorimotor level and high-level cognition" (Guerin's comments to (Perotto 2013, p. 312). Appendix A lists topics, keywords, and references from Cangelosi and Schlesinger's (2015) excellent book on developmental robotics. Other AI constructivism surveys are available by Lungarella et al. (2003), Asada et al. (2009), and Ziemke (2001).

## 2 The Guerin 2008 Survey

Guerin (2008a) summarizes some history and key developments of AI constructivism beginning with Drescher (1989/1991) and continuing through Cohen PR et al. (1997), Rosenstein et al. (1997), Stojanov et al. (1997, 2001), Chaput (2004), Barto et al. (2004), Barto and Mahadevan (2003), Bakker and Schmidhuber (2004), Holmes and Isbell (2006), Perotto and Alvares (2006), St. Amant et al. (2006), Chang et al. (2006), Oudeyer et al. (2007), Bondu and Lemaire (2007), Konidaris and Barto (2007), Lee et al. (2007), Stoytchev (2007), and Kuipers and colleagues (e.g., Pierce and Kuipers (1997), Kuipers (2000), Kuipers and Beeson (2002), Modayil and Kuipers (2004), Kuipers et al. (2006), Provost et al. (2006), Olsson et al. (2006), Modayil and Kuipers (2007), Provost et al. (2007)). This section briefly summarizes these works—characterizing the components, algorithms, and architectures in terms of Piaget's schemas, adaptation processes, and staged development.

For Drescher (1989/1991), the schema is a simple Context-Action-Result[9] structure, e.g., SeeHandAtPositionX-MoveHandBackward-SeeHandAtPositionY. These schemas can grow to represent more sophisticated knowledge via mechanisms of chaining, synthetic item generation, spinoff schemas, and composite actions. The general learning algorithm, the "schema mechanism," utilizes processes of induction (i.e., marginal attribution), abstraction, and invention. It is focused on the first 5 sub-stages of Piaget's Sensorimotor stage, while presumed to apply to the later stages as well.

In Cohen PR et al.'s (1997) Neo system, temporally-coincident sensations and percepts form schemas called *fluents* which combine into increasingly sophisticated structures called composite fluents, chains, classes, and physical schemas. Base fluents are "(attribute value)" pairs where the attribute identifies a sensorimotor stream and the value is a particular sensation or action relevant to that stream, e.g., (sight-shape rattle-shape), (sight-color red), (tactile-hand wood), (hand close), (voice cry).

[9] Context-Action-Result (or CurrentState-Action-NextState) is a common design pattern for constructivist learning (e.g., Stojanov, 2001; Bakker and Schmidhuber, 2004; Perotto and Alvares, 2006). It may also be seen to be analogous to Bayesian reasoning where Prior (context) + New Evidence (action) = Posterior (result) and to rules in production systems (Klahr and Wallace, 1976).

Rosenstein et al. (1997) extend Cohen and attempt to learn schemas as Lakoff and Johnson-style *image schemas* (Lakoff and Johnson 1980; Johnson M 1987[10]; Gibbs and Colston 1995) represented as 2-dimensional activity maps.

Stojanov (2001) describes aspects of the Petitagé system including *expectancies* having a SensorState-Action-SensorState structure and schemas that are sequences of actions of arbitrary length.

Chaput (2004) presents a *Constructivist Learning Architecture* that uses Self-Organizing Maps (Kohonen 2013) representing correlations of environmental features as nodes in a hierarchy of increasingly higher-level knowledge. He summarizes six *Information Processing Principles* from Cohen LB et al. (2002) as key to "a domain-general learning system that provides continuous learning from the environment" which characterizes staged learning as:

1) Infants are endowed with an innate information-processing system,
2) Infants form higher schemas from lower schemas,
3) Higher schemas serve as components for still-higher schemas,
4) There is a bias to process using highest-formed schemas,
5) If, for some reason, higher schemas are not available, lower-level schemas are utilized,
6) This learning system applies throughout development and across domains. (Chaput 2004, pp. 9–10)

Barto et al. (2004) explore *intrinsically motivated learning* as a kind of reinforcement learning (RL) that can drive a developmental agent to learn increasingly complex behaviors. They focus on a surprise/novelty measure as an intrinsic reward signal. Agents repeatedly perform actions that elicit such rewards and lose interest as the novelty/reward subsequently decays. Barto et al. (2004) suggest such a policy can learn closed-loop control rules called *options*, which are a kind of schema consisting of an *option policy*, *initiation set*, and a *termination condition*. Details on the options framework, which combines RL with Markov Decision Processes (MDPs) and semi-Markov decision Processes (SMDPs), are provided by Sutton et al. (1999). Sutton (1999) wrote, "If the right features are represented prominently□, then learning is easy□; otherwise it is hard.□ It is time to consider seriously how features and other structures can be constructed automatically by machines rather than by people. … In psychology, the idea of a developing mind actively creating its representations of the world is called constructivism. *My prediction is that for the next tens of years RL will be focused on constructivism*."

In an earlier paper, Barto and Mahadevan (2003) describe work similar to the *options framework* that also involves hierarchical RL and SMDPs, i.e., Hierarchies of Abstract Machines (HAMs) by Parr and Russell, and MAXQ Value Decomposition by Dietterich. HAMs coordinate the execution of multiple finite-state machines to achieve goals such as in an exemplary robot navigation environment where higher-level behaviors are defined such as *back-off* or *follow-wall*. Regarding MAXQ, Barto and Mahadevan (2003) state: "Unlike options and HAMs, … the MAXQ approach does not rely directly on reducing the entire problem to a single SMDP. Instead, a hierarchy of SMDPs is created whose solutions can be learned simultaneously." MAXQ structures a hierarchy of actions (subtasks) in a *task graph* with nodes representing actions like *Get*, *Put*, *Pickup*, *Dropoff*, *Navigate*, and move *North*, *South*, *East*, and *West*.

[10] Johnson M (1987, p. xiv) defines image schema as "a recurring, dynamic pattern of our perceptual interactions and motor programs that gives coherence and structure to our experience." It was called *image schema* because it "functions somewhat like the abstract structure of an image" (ibid, p. 2).

Bakker and Schmidhuber (2004) describe the HASSLE algorithm (Hierarchical Assignment of Subgoals to Subpolicies LEarning algorithm)—a hierarchical reinforcement learning algorithm that uses intrinsically motivated learning to allow high-level policies to discover sub-goals by clustering sensor data and have low-level policies specialize on reaching sub-goals. Observation vectors are clustered using ARAVQ (Adaptive Resource Allocation Vector Quantization). High-level and low-level options[11] (schemas) are learned as high- and low-level value functions using temporal difference learning at different temporal resolutions. The system claims to have automatically achieved three levels of learning for an agent navigating a simulated office grid world.

Holmes and Isbell (2006) employ looping Prediction Suffix Trees to represent deterministic Partially Observable Markov Decision Processes to identify what state an agent is in and predict what can happen next.

Perotto and Alvares' (2006) schemas consist of Context-Action-Expectation triples where each element is represented by a vector of sensor (or effector) states where each state can have a value of true, false, or undefined. The schemas are organized into trees where root nodes represent general situations and more specific situations are encoded in the more distal branches and leaves. In Perotto et al. (2007), a Constructivist Anticipatory Learning Mechanism (CALM) is introduced that extends Context-Action-Expectation schemas to more complex (partially deterministic and partially observable) environments by using synthetic elements to represent abstract or hidden properties.

St. Amant et al. (2006) extends Rosenstein et al. (1997) in describing an *image schema language* (ISL) for representing Lakoff and Johnson-style *image schemas* (Lakoff and Johnson 1980; Johnson M 1987; Gibbs and Colston 1995). Therein, "image schemas are objects, as in the object-oriented data model" where "each schema has a set of operations that determine its capabilities" and "internal slots … that permit image schemas to be related to each other through their slot values." Sets of image schema instances can be used to define each state in a state machine[12]. St. Amant et al. (2006) also define learned sequences of image sequences called "gists" which serve as a kind of higher-level reusable action structure tied to particular goals.

Chang et al.'s (2006) Jean system builds on St. Amant et al. (2006) using ISL concepts to implement an infant-like agent that attempts to learn, execute, and extend schemas in a simulated playpen environment. It introduces an Experimental State Splitting (ESS) algorithm for building a world model by composing schemas into *gists* (composite schemas) and *differentiating* (splitting) existing states into new states. The authors claim their policies allow Jean to learn causal relationships. Cohen PR et al. (2007) provides more details on action schemas and Jean's ability to learn and repurpose (transfer) gists in a 3-D real time strategy game environment.

Oudeyer et al. (2007) use vector exemplars representing sensorimotor experiences, which are partitioned into distinct regions using an Intelligent Adaptive Curiosity (IAC) algorithm that utilizes intrinsically motivated learning similarly to Barto et al. (2004). Each region is associated with a particular learning machine, e.g. neural network, support vector machine, or Bayesian machine. The number of exemplars allocated to a region is limited to 250. Each region maintains a list of error rates representing how much the results of executed actions vary from what was predicted. These lists are subsequently used to select the best actions for new situations. Bondu and Lemaire (2007) extend this work by reformulating the

[11] Bakker and Schmidhuber use the notation: $o_s$ for start state observations, $o_g$ for goal state observations, $\pi$ for policies (action sequences), and $r$ for rewards, each with an L or H superscript indicating whether they are low-level or high-level.
[12] The ISL description in St. Amant et al. (2006) specifies three types of interacting schemas: static, dynamic, and action schemas. It appears static schemas are object-oriented-like objects. Dynamic schemas involve dynamic maps as sketched in Rosenstein et al. (1997). Action schemas add the state machine representation aspect.

IAC framework using active learning[13] terminology. They propose an improved metric for region splitting that balances the mixture rate and relative density of exemplars.

Konidaris and Barto (2007) extend the work of Barto et al. (2004) by learning schemas in the options framework wherein *agent-space options* are generated in addition to the previously described *problem-space options*. Agent-space skills are reportedly transferrable to new tasks that have different problem-spaces.

Lee et al. (2007) describe a Lift-Constraint, Act, Saturate (LCAS) approach wherein learning consists of an agent performing actions in a constrained fashion until no new actions are possible. When learning saturates at the current level, constraints limiting the range of actions are lifted and learning continues to the next level. Lee et al. (2007) suggest levels of constraints may occur in an order such as gross motor coordination of limbs, followed by proprioception, tactile sensing, auditory sensation, visual sensation, and fine motor control. They explored primitive motor and sensory variables utilizing a "schema" consisting of 2-D Sensory Space and Motor Drive Space maps.

Stoytchev (2007) describes a developmental approach a robot can use to autonomously learn the capabilities of its own body and extend those to tool use. Learning starts with random motor babbling. A *Robot Body Schema* (RBS) that builds on the *Self-Organizing Body Schema* (SO-BoS) of Morasso and Sanguineti (1995) is used to situate on-going development. The SO-BoS defines a multiplicity of processing elements ($PE_i$) that each learn prototypical (preferred) *body icons* which consist of a vector of motor features $\theta_i$ (e.g., joint positions) and associated sensory features $V_i$ (e.g., visual locations): ($\theta_i$, $V_i$). Adjacent PEs learn similar body icons which can be activated in parallel. Stoytchev (2007) also describes how different *body frames* can be learned by clustering body markers based on co-movement patterns.

Kuipers and colleagues (see references above) focused on how to bootstrap learning about an agent's spatial environment. They describe methods an agent can use to learn a model of its sensors, a model of its motor apparatus, and a set of behaviors that allow it to abstract its environment to a discrete representation of places and paths. They propose a *spatial semantic hierarchy* (SSH) having egocentric models of the agent's sensorimotor apparatus at the bottom level and a representation of the environment defined by a discrete set of views and actions at the top. The SSH consists of five levels: sensorimotor, control, causal, topological, and metrical. It is constructed by learning sensorimotor features of the agent/environment and then proceeds to learning actions that predict sensorimotor events, to doing further (causal) abstraction into a finite set of views and actions, to learning global representation of world structure to, finally, filling in details about world structure. This work is notable for its degree of constructivism even at the lowest levels, e.g., sensory features are learned by *feature generators* that detect similar inputs over time and have similar frequency distributions. Motion control features are similarly learned via general low-level algorithms. At the causal level, Kuiper's schemas represent state transitions that similar to Drescher (1987/1991) and others, i.e., View-Action-View. This work also attempts to consolidate ideas from Sutton et al. (1999) (i.e., options), Chaput (2004) (i.e., Kohonen Self-Organizing Maps), and others into the context of the SSH architecture.

Olsson et al.'s (2006) extensions to low level sensorimotor learning in the SSH framework use information theoretical metrics for correlating sensors and developing *sensoritopic maps* that reflect the informational geometry of the agent's sensors. The maps are used to generate *sensorimotor laws* that specify how agent actions affect its sensors.

Modayil and Kuipers (2004, 2007) go beyond the SSH spatial learning model to describe unsupervised learning of objects via steps of Individuation, Tracking, Image Description, and Categorization. Objects are described by suitably constituted sensor readings that are clustered to identify particular objects. They describe an *object ontology* by a tuple "(Trackers, Perceptual Functions, Concepts, and Actions)." Object concepts are formed by clustering object percepts that are stable in time. Associated actions are actions that have been learned that reliably change a particular object's percepts.

[13] In active learning, the *most informative* training examples are sought (and generally labeled by an expert) to accelerate model learning.

## 3 Subsequent Work by the Guerin Referents

This section summarizes constructivist AI work since 2008 by those surveyed in (Guerin 2008a). People not mentioned, like Drescher, Chaput, Bakker, Holmes, and Isbell, appear to have not continued this line of development.

Guerin and colleagues have published on learning object manipulation and tool use (Guerin et al. 2013) and on automatically identifying tools and their functions (Abelha et al. 2016; Abelha and Guerin 2017). Guerin et al. (2013) provides a summary of constructivist developmental paths, methods, and potential knowledge representations consistent with developmental psychology. Celikkanat et al. (2015a) describe work on modeling the *context* of an agent using a Random Markov Field "concept web" as a latent variable of an incremental Latent Dirichlet Allocation (LDA) process. They define context as the "set of active concepts in the scene" and leverage context for agent cognition and behavior[14].

Subsequent to Neo, Jean, ISL, Wubble World[15], and related systems described in (Cohen PR et al. 1997; Rosenstein et al. 1997; St. Amant et al. 2006; Chang et al. 2006; and Kerr et al. 2007), Paul Cohen and colleagues continued work in the AI constructivist vein on language learning (Hewlett and Cohen 2009; Hewlett 2011), an Action Schema Generator that can acquire and represent semantic knowledge based on a simulated agent's experience (Mu 2009), representing activities as finite state machines (Kerr et al. 2011), and ways for humans to teach autonomous agents (Kaochar et al. 2011). More recently Cohen has explored leveraging human-machine synergies, where machines focus on mining and synthesizing data into integrated causal models and "pushing" results to humans (Cohen PR 2015; Cohen PR 2018; Cohen PR 2020). This approach advocates AIs be deployed as knowledge extraction and synthesis tools rather than agents that learn and reason like humans. The Big Mechanism program (Cohen PR 2015) uses AI to automatically build causal models by extracting causal claims from scientific papers. Associated DARPA-affiliated programs "demonstrated that machines can read text, tables, equations and even FORTRAN code from legacy models, and build comprehensive models of the world's complicated, interacting systems" (Cohen PR 2020). Cohen contrasts this approach with typical big data and machine learning methods that focus on learning patterns (correlations) and not on discovering causal mechanisms.

Kuipers and colleagues continued work on autonomous environmental mapping, including further development of the Spatial Semantic Hierarchy (SSH) model through variants called hybrid-SSH (HSSH) (Kuipers 2008) and Hierarchical Hybrid Spatial Semantic Hierarchy ($H^2$SSH) (Johnson CE 2018). The HSSH "represents a robot's environment using four distinct layers that provide metrical and topological representations of both small-scale and large-scale space." $H^2$SSH (Johnson CE 2018) "improves on the HSSH by providing hierarchical representations of both local and global space that improve the scalability of the topological mapping problem." $H^2$SSH updated the HSSH to represent three topological features: path segments, decision points, and destinations; support richer semantics for path segments; and allow nesting of topological maps to support more complex spatial environments. Prior to the $H^2$SSH work, Mugan (2010) described an algorithm called the Qualitative Learner of Actions and Perception (QLAP) that autonomously learns predictive models of an environment and a set of hierarchical actions suitable for acting on it. Knowledge is represented as a network of plans and actions.[16] In other work, Xu C (2011) introduces an Object Semantic Hierarchy (OSH) that constructively recognizes and models objects in the environment. As summarized therein: "The agent initially treats everything in the sensory stream as noise. By repeatedly identifying new invariants to reduce the noise, the agent progressively builds models for the background world and foreground objects. For the background world or each foreground object, the model evolves from 2D2D to 2D3D to 3D3D." In another thread, Liu et al. (2011) explore how high-level semantic action concepts/attributes, both manually defined and autonomously learned, can be used to represent higher-level human actions. For example, they note attributes of *single-leg-motion*, *arm-over-shoulder-motion*, and *torso-up-down-motion* may effectively classify the action of *golf-swinging*. Thus they explore potentially useful decompositions of verbal/action concepts as opposed to the more-studied object recognition and classification domains. Mittelman et al. (2014) describe an attribute tree process (ATP) that learns a tree hierarchy of semantic concepts using an unsupervised Bayesian method. They provide evidence that ATP may be superior to agglomerative hierarchical clustering (AHC) and the factored Bernoulli likelihood model (FBLM) for doing concept clustering.

[14] More formally, they note the probability of a concept being applicable to a scene is $P(c)* = \sigma \times P(c) + (1-\sigma) \times P(c|\chi)$, where $c \in C = N \cup A \cup V$ is a concept (consisting of Noun, Adjective, and Verb parts), P(c) is the MRF-decided probability of the concept c, $\chi$ is the context, $P(c|\chi)$ is the probability of the concept given the context (decided by Incremental-LDA), and $P(c)*$ is the updated value of the concept probability. $\sigma$ is a hyper-parameter used for regulating the strength of contextual feedback.
[15] Wubble World focused on constructivist language development.

Stojanov (2009) gives a brief summary of AI and robotics work influenced by Piaget from years 1963 through 2008. Stojanov and Indurkhya (2013) discuss roles of perceptual similarity and analogy in a constructivist view on creativity. They see analogy as a "core mechanism in human cognitive development rather than a special skill among many" (in distinction to Piaget) and they discern perceptual similarity as a key to creative problem solving.

Perotto (2013) reprises and refines CALM (Perotto et al. 2007), adding a meta-architecture called the coupled agent environment system (CAES) in which an intrinsically motivated CALM agent is situated. The CAES consists of an agent (A) logically comprised of separate body (B) and mind (M) components that exchange percepts (p) and control signals (c). It also defines a "world outside the mind" (W) made up of the agent's body (B) and the wider environment (E) that interact through situations (s) and actuations (a). Perotto notes CAES and a CALM agent interact as two mutually dependent *dynamical systems*. The world is modeled as a factored and partially observable Markovian decision process (FPOMDP). The paper is followed by commentary by Guerin, Butz, Thorisson, Stojanov, Bickhard, Degris, and Scott on CALM, CAES, and the state of the art of constructivist solutions in general—providing a useful snapshot of the status of constructivist AI development in 2013. They suggest the main unsolved challenge is the ability to build up cognitive capabilities from the sensorimotor level to higher conceptual levels in the presence of complex, high-bandwidth environmental sensations and noise, e.g., create symbols, abstract structures, and models.

Oudeyer continued work on constructivist AI "focusing on sensorimotor development, language acquisition and life-long learning in robots."[17] In response to Lake et al. (2016), Oudeyer (2017a) discusses additional "crucial ingredients" necessary for autonomous learning: curiosity and intrinsic motivation, social learning and natural interaction with peers, and embodiment. In (Oudeyer 2017b), AI constructivism (development) is described as generating a "complex dynamical system, characterized by spontaneous self-organization or emergent patterns at multiple scales of time and space." In (Santucci et al. 2020), Oudeyer and colleagues summarize work on intrinsically motivated open-ended learning in autonomous robots and highlight key open problems including autonomous generation of goals, learning policies to achieve goals, appropriate use of intrinsic motivation ("to support learning compact representations of environment states"), and a need for better ways to encode goals and skills.

[16] QLAP represents environmental models using dynamic Bayesian networks (DBNs) and represents action plans using Sutton et al.'s (1999) options framework. Many small Markov Decision Processes (MDPs) are created to represent small aspects of the environment.
[17] http://www.pyoudeyer.com/bio/.

Barto, Sutton, and colleagues[18] continued exploring sensorimotor approaches to knowledge acquisition (Sutton 2012) building on their work in reinforcement learning. Sutton (2009) summarizes status of the Predictive Empirical Abstract Knowledge (PEAK) project: "an attempt to understand world knowledge in terms of a minimal ontology of sensorimotor experience" where they endeavored to "connect low-level signals [consisting of sensations and actions] to higher-level representations in such a way that the knowledge remains grounded and autonomously verifiable." The approach, which included temporally abstract *options*, option models, and temporal-difference (TD) networks, were evaluated in two simulated environments (bit-to-bit world and compass world) and in a sensor-rich robotic environment (Critterbot). Sutton et al. (2011) describe the *Horde* architecture, which improved upon TD-networks and options using gradient-TD methods and *General Value Functions* (GVFs) as knowledge representations (schemas). The architecture relies on "a large number of independent reinforcement learning sub-agents, or *demons*" that learn in parallel, each "responsible for answering a single predictive or goal-oriented question about the world. … Each demon has its own policy, reward function, termination function, and terminal-reward function." Barto et al. (2013) summarize work related to exploring and representing behavioral hierarchies. They describe refinements to the Sutton et al. (1999) options framework while focusing on using hierarchical reinforcement learning (HRL) to learn hierarchies of behavioral modules. Da Silva et al. (2014) introduce an active learning method for efficiently acquiring *parameterized skills* that are more granular and reusable than skills originally defined in the options framework. Niekum et al. (2014) propose constructivist improvements in robotic *learning from demonstration* (LfD) by discerning "repeated structure at multiple levels of abstraction in demonstration data," thus discovering semantic knowledge about the world and building up a "library of skills" over time. Sutton and colleagues worked on improving the performance of online reinforcement learning algorithms to "true" online versions. These include *true online TD(λ)*[19] (van Seijen and Sutton 2014), *true online GTD(λ)* (van Hasselt et al. 2014), *true online Sarsa(λ)* (van Seijen and Sutton 2014), and *true online emphatic TD(λ)*[20] (Sutton 2015). De Asis et al. (2020) suggest *fixed-horizon temporal difference* (FHTD) methods, which predict "the sum of rewards over a fixed number of future time steps," as being a further advance. Veeriah et al. (2017) describe the *crossprop* neural network-learning algorithm, which unlike backprop, can "learn to reuse the learned features for solving new and unseen tasks."

Stoytchev has since collaborated on work to learn multi-sensory features to categorize common objects (Sinapov et al. 2014) and on work to teach robots to do other object recognition and manipulation tasks[21].

Schmidhuber (2010) summarizes his work on *intrinsic motivation* from 1990 to 2002. He suggests the purpose of intrinsic motivation is to "provoke event sequences exhibiting *previously unknown but learnable* algorithmic regularities" and to build an agent "that never stops generating non-trivial & novel & surprising data." Key components of such agents are: "an adaptive world model …, a learning algorithm that continually improves the model …, *intrinsic rewards* measuring the model's improvement …, [and] a separate reward optimizer or reinforcement learner." Intrinsic rewards assign values to "the discovery or creation of *novel patterns,*" which Schmidhuber equates with "fun or internal joy." He reviews his previous "practical but non-optimal" proposals where reward values are: (1) proportional to the model's prediction errors[22], (2) proportional to expected improvement (first derivative) of prediction error, (3) proportional to relative entropies of learning agent's priors and posteriors, and (4) determined through zero-sum games between a "right brain" and "left brain." Schmidhuber (2013) describes PowerPlay—an "implementation of basic principles of creativity" inspired by playful behavior in animals[23]. PowerPlay evaluates new tasks (problems) against existing skills (*problem solvers*) and, if a task cannot be *solved* by an existing skill, it modifies existing skills and remembers one that solves the new task and all previous related tasks. Thus solvers (skills, schemas) become more general over time. Types of solvers considered include deterministic universal computers, finite state automata, and feedforward neural networks. Procedures for task invention, solver modification, and correctness demonstration are described. Schmidhuber (2015) describes recurrent neural network-based AIs (RNNAIs) that learn to think by combining an RL *controller* with an RNN-based *predictive world model*. Schmidhuber (2017) suggests technology like Long Short-Term Memory (LSTM)[24] combined with artificial curiosity and creativity will soon enable "an AI that incrementally learns to become as smart as a little animal—curiously and creatively and continually learning to plan and reason and decompose a wide variety of problems into quickly solvable (or already solved) sub-problems."[25] Ha and Schmidhuber (2018) present a simplified framework based on Schmidhuber's work from 1990 – 2015 that combines a large RNN-based world model with smaller controller models to enable agents to learn compact and simple task completion policies. They use a Variational Autoencoder to encode image frames, a Mixture Density Network combined with an RNN to predict future frames, and a simple linear model to control outputs. Van Steenkiste et al. (2018) describe a *relational neural expectation maximization* (R-NEM) method for learning objects and their physical interactions in an unsupervised manner from raw video images. Greff et al. (2020) propose a unifying framework aimed at solving the *binding problem* and approaching human-level generalization by segregating sensory inputs, maintaining separate symbolic (object) representations, and using those representations to compose new inferences, predictions, and behaviors. They provide an extensive survey of the *object representation* literature spanning AI and cognitive psychology. Csordás et al. (2021) propose modifying the Transformer architecture by adding a *copy gate* and using *geometric attention* to allow Transformers to learn more generally applicable rules.

---

[18] Links to Barto's and Sutton's publications are at https://people.cs.umass.edu/~barto/pubs-Barto.html and http://incompleteideas.net/publications.html#beyond_reward.

[19] This is a temporal difference algorithm that "combines basic TD learning with eligibility traces to further speed learning."

[20] Emphatic approaches generate knowledge structures by selectively emphasizing or de-emphasizing updates on different time steps. Gu et al. (2019) suggest emphatic TD may be superior to other reinforcement learning algorithms.

[21] See publications list at http://www.ece.iastate.edu/~alexs/lab/publications/index.html.

[22] Schmidhuber (1991a) suggests intrinsic reinforcement be calculated based on the difference between predicted and actual experience where "zero reinforcement should be given in case of perfect matches, high reinforcement should be given in case of 'near-misses', and low reinforcement again should be given in case of strong mismatches." These cases correspond to concepts that are already learned (boring), effectively learnable (curiosity piquing), and unlearnable (befuddling).

## 4 Other Constructivists

Early purveyors of constructivist ideas, appropriately not surveyed in (Guerin 2008a) given their less direct influence on AI constructivism, have been discussed by Ernst von Glasersfeld (1984, 1990), Michael Arbib (1992), D. C. Phillips (1995), and Sandra Marshall (1995). Perhaps most influential were John Locke (1690) for his embrace of empiricism, rejection of innate ideas, and promotion of the principle that sensations give rise to secondary qualities that enable all ideas[26]; Immanuel Kant (1781) for challenging Locke's notions of substances and primary qualities and promoting the idea that all knowledge is derived from sensations[27]; William James for his radical empiricism[28] and proposing how "blooming, buzzing confusion" gives way to knowledge through learning objects and relations via discrimination and synthesis (James 1890)[29]; and Lev Vygotsky (2012) for emphasizing the significance of social interactions for creating schemas.

[23] Playful behaviors are tasks that are usually self-generated and used to acquire skills that can be applied later to tasks encountered in the environment.
[24] A type of memory cell used in hidden units of Recurrent Neural Nets (RNNs) that effectively handles long-term temporal dependencies (Greff et al. 2017).
[25] Such a goal is being pursued by NNAISENSE, a company Schmidhuber cofounded with Gomez, Koutnik, Steunebrink, and Masci. (https://nnaisense.com). Their current focus is on intelligent automation, deep learning, and AGI.
[26] Locke's *primary qualities*, interpreted as objective "measurable aspects of physical reality" (Wikipedia, Primary/secondary quality distinction), are unnecessary from a constructivist perspective. *Secondary qualities* interpreted as subjective qualities perceived by the senses, on the other hand, are fundamental to constructivism.
[27] Agents have access only to the *phenomenal world*. The *noumenal world* (things as they really are) is inaccessible. Reason is used to create additional knowledge from sensed experiences. Kant's conceptions of *schema*, characterized as rules or filters mapping sensations to concepts, map well to Piaget and other modern psychologists (Scaglia 2018; Marshall 2015).
[28] Radical empiricism posits that *all* knowledge, i.e., objects and relations, comes from direct experience (James 1904).
[29] James uses the terms "association and dissociation," "subdividing and uniting," and "break asunder and reunite." He also notes the importance of selective attention and sensory filtering: "We do far more than emphasize things, and unite some, and keep others apart. We actually ignore most of the things before us" (James 1890, p. 284).

This section does not examine such influences. Rather, it considers works subsequent to Piaget (1934/1954) that are not in (Guerin 2008a) and more directly relevant to AI constructivist implementations. As in previous sections, the work is presented chronologically.

George Kelly's (1955) Personal Construct Theory (PCT) defines *constructs* (schemas) as "transparent patterns or templates created by humans and lower animals that fit over the realities of the world and enable them to chart a course of behavior."[30] Developed in the context of clinical psychology, PCT posits that agents construct and adapt personal mental models that allow them to effectively predict events. Agents function as naïve scientists that form and test hypotheses and channel their thoughts and behaviors consistent with plausible hypotheses.[31] Different, often incompatible, constructs are used to make sense of events over time—a process Kelly called *constructive alternativism*. A distinguishing feature of PCT is that mental constructs consist of *pairs of mutually exclusive binary values* (e.g., old-young, good-bad, happy-sad, light-dark, comes_when_I_cry, doesn't_come_when_I_cry). These *dichotomous constructs* define axes of a multi-dimensional *psychological space* with elements of experience representing points in the space (Kelly 1955, Ch. 6; Gaines and Shaw 2012, p. 21). Constructs form hierarchies, where higher level constructs may change by "invoke[ing] new arrangements among the systems which are subordinate to them" (Kelly 1955, p. 55). To operationalize the theory, Kelly introduced a cognitive mapping technique called *repertory grids* that provides methods for eliciting constructs and observational elements from agents and organizing them into rows and columns in a matrix (Shaw 1978; Gaines and Shaw 1991; Curtis et al. 2008). This data can be clustered or otherwise compared to progressively refine constructs and better understand agent's mental models. Proponents have nudged PCT in constructivist AI directions by building computer applications that automate the creation and analysis of repertory grids, support the use of repertory grids in non-psychological applications, facilitate knowledge acquisition for expert systems, and aid in constructing semantic networks (Shaw 1978; Gaines and Shaw 1991; Curtis et al. 2008; Gaines and Shaw 2012).[32] Boeree (2006) provides a summary of Kelly's life and works.

Ceccato and Zonta (1961) present work on machine translation (MT)[33] that leveraged constructivist principles from their Italian Operational School of philosophy (Hutchins 1986). They specify four key operations (adaptation mechanisms) for constructing thoughts: *differentiation*, *figuration*, *categorization*, and *correlation*. Differentiation refers to detecting changes of state, which gives rise to distinctions like dark-light, hot-cold, resistant-yielding, green-red-yellow, and silence-noise. Figuration detects changes of place, which allows recognition of shapes and volumes. Categorization corresponds to mental classification based on temporal differentiation, which "gives us the mental, or logical, categories, including, for example, substance, accident, subject, object, and, or, with, also, by, state, point, line, surface..." Correlation, which represents thought itself, links items created by the other three operations into distinct temporal units (Ceccato and Zonta 1961; Ceccato 1962, pp. 62 - 65; Hutchins 1986). Ceccato's MT implementation reflected these operations via *correlational tables* and *correlational nets*. The latter consisted of a network of *correlational triads* (the temporal units) each unit comprising a *correlator*[34] and first and second *correlatums*. Semantic net-like constructs called *notional spheres* and frame-like constructs called *constellations* were used to resolve ambiguities (i.e., polysemanticity) in MT phrase mapping (Ceccato 1962, pp. 83 – 87). Ceccato (1967) mentions early work on what he called "The

[30] Slightly paraphrased from Kelly (1955, p. 7, pgphs. 1 – 2).

[31] Kelly's *fundamental postulate* states: "A person's processes are psychologically channelized by the ways in which he anticipates events." The fundamental postulate is supplemented with 11 corollaries (Kelly 1955, Ch.2). See Appendix I in (Curtis et al. 2008) for a concise summary.

[32] Gaines and Shaw (2012, Sect. 4.3) note similar construct descriptions arising in various academic areas, including Bartlett's *schema* (human memory processes), Ranganathan's *faceted taxonomy* (library science), Kelly's *conceptual/repertory grid* (personal construct psychology), Minsky's *frames* (knowledge representation), Piaget's *schema* (developmental psychology), Filmore's *frames* (linguistics), and Barsalou's *frames* (cognitive psychology).

[33] For Ceccato et al., MT signified "*mechanical* translation."

[34] A modest number of primitive correlators (i.e., 100 - 200) were envisioned. They consisted of "the conjunctions and prepositions, punctuation marks, and relations such as subject-predicate, substance-accident (i.e., noun-adjective), apposition, development-modality (i.e. verb-adverb), and comparison" (Hutchins 1986). Correlators temporally link correlatums.

Fourth Approach to MT," aimed at producing a proof-of-concept machine "which observes and describes the events of its surroundings through the actual operations of perception, representation, categorization, etc."

Ayn Rand's *Objectivist epistemology* (Rand 1967) is notable for its constructivist principles. Claims include: all knowledge is based on perception; a conceptual hierarchy spans from *sensations* to *perceptions* to *concepts*; understanding infant development is key to understanding concept formation; developmentally, existing things (*existents*) are recognized progressively—first as entities, then *identities*, then *units* (i.e., instances); concepts are constructed by *differentiating* (i.e., abstracting out) *and integrating* (i.e., uniting) attributes from two or more units using "in large part, a mathematical process;" units are distinguished as having shared attributes with different values; *conceptual common denominators* such as length, shape, kinds of motion, and color are used to build higher-level concepts; concepts are structured in a hierarchy of increasing abstraction. Boydstun (2012) discusses similarities between Rand and Piaget's ideas. He suggests Rand may have adapted her epistemology to more closely track Piaget given her familiarity with Flavell's (1963) account of Piaget.

Michael Cunningham (1972) proposes a constructivist system that seeks to synthesize Hebb's neural assemblies, Piaget's schemas, and Sokolov's cognitive reflexes. It focuses on modeling mechanisms that are active during the first two years of human development—corresponding to Piaget's Sensorimotor sub-stages.[35] Building-block *elements*, intended to approximate Piaget schemas and Hebb cell assemblies, are defined as neural structures that reliably produce a particular spatial and temporal firing pattern given a corresponding pattern of stimulation. The key dynamic of the system is characterized as "an ever continuing, ever changing and expanding *circular reaction* with the environment." Circular reactions are visualized as loops connecting *input elements* through *reflex links* to *output elements* to objects in the environment and back to input elements. The basic schema structure consists of an input element linked to an output element where each represents activations of particular sensory receptors and motor effectors respectively. Activated elements are said to *reverberate*. Initial circular reactions are entirely instinctual and reflexive[36]. Over time, as a result of experience, additional elements form, become more interlinked, and may exist independently from input and output elements ("tight little knots of mutually interconnected assemblies") enabling more complex behaviors and cognitive capabilities (e.g., memories, chains of thoughts, separation of goals from means). The main mechanism for constructing more complex elements is the linking together of lower-level elements that reverberate at the same time.[37] Drescher (1989, p. 201) cites Cunningham (1972) as the inspiration for his own work. Cunningham and Gray (1974) present an implementation based on (Cunningham 1972). They describe a computer simulation of an infant vocal tract capable of generating outputs based on sensations from rudimentary audio, kinesthetic, and proprioceptive inputs. They claim their system achieved learning spanning Piaget's first three sensorimotor sub-stages and speculate the model could be adapted to achieve sub-stage four.

Michael Arbib embraces constructivist principles in his computational neuroscience and AI research. In (Arbib 1972) he writes, "The animal perceives its environment to the extent that it is prepared to interact with it. … Perception of an object generally involves the gaining of access to 'programs' for controlling interaction with the object, rather than simply generating a 'name' for the object." He argues that brain models and AI implementations should be distributed, action-oriented, and layered, and such models consist of perceptual and motor programs. Arbib's (1981) analysis of visuomotor coordination in frogs

[35] Cunningham, closely following Piaget, summarized the 6 sub-stages as: (I) reflex exercise, beginning from birth, (II) primary circular reactions, beginning in second week, (III) secondary circular reactions, beginning in fourth month, (IV) familiar procedures in new situations, beginning in eight month, (V) active experimentation, beginning in eleventh month, (VI) mental recombinations, beginning in second year.

[36] Circular reactions are like central pattern generators (CPGs), i.e., "biological neural circuits that produce rhythmic outputs in the absence of rhythmic input" (Wikipedia, https://en.wikipedia.org/wiki/Central_pattern_generator) and innate behavior patterns (Witkowski 1997).

[37] Cunningham gives an example of a hand sucking behavior learned as a conjunction of a previously learned arm flexion element occurring simultaneously (initially by chance) with a previous sucking reflex element.

(*Rana computatrix*) motivates a proposal for *assemblages of perceptual and motor schemas* to constitute an "animal's internal representation of the world." Arbib and Hesse's (1986) "The Construction of Reality" influenced Arbib's view of cognitive models as EvoDevoSocio constructions (Arbib 2018), therein formulating *social schemas* as *constructed realities* that encode mental building blocks of *societies*. [38] Arbib (1989) updates (Arbib 1972) with "programs" rechristened as "schemas" and *schema theory* described in more detail. Arbib (1992) summarizes schema theory—noting historical precedents, key architectural elements, and experimental findings. Arbib (2018) provides more information about schema theory, including a discussion of Draper et al.'s (1989) schema implementation for the VISIONS system and collaborations with Corbacho on *schema-based learning*. Corbacho (1997) describes schema-learning mechanisms of *tuning*, *construction*, and *active learning*. Weitzenfeld et al. (1998) present a neural-based schema architecture that builds on Arbib's work and utilizes an *abstract schema language* to integrate neural networks into schemas. Corbacho (2019) specifies a *self-constructive AI* framework that constructs *predictive schemas*, *dual (inverse model) schemas*, and *goal schemas*—each embodying a different behavior and communicating using input ports, output ports, and variable mappings. Arbib (2021) crystallizes his earlier work summarizing schemas as *interacting functional units* organized as networks of *basic motor schemas* and *perceptual schemas* that are instantiated as *multiple instances* and activated *cooperatively* using *bottom-up* (*data-driven*) and *top-down* (*task-driven*) signals to elicit appropriate behaviors (e.g., object recognition, motor actions). Explicit executive control is not needed. New schemas may form as assemblages of old schemas, be adaptively tuned, and become primitives.

Joseph Becker (1973) details the JCM model for encoding experiential information. He provides an example of a schema[39] as [Sensation$_1$ -> Action$_1$ -> Action$_2$ => Sensation$_2$] where if Sensation$_1$ is sensed, Action$_1$ and Action$_2$ may be executed in sequence to elicit Sensation$_2$. More generally, conditions (contexts) and actions on the left side of the Big Arrow (=>) result in (predict) sensations on the right side. Sensation and action elements, called *kernels*, are n-tuples of nodes representing features of the environment. The kernels on the left side of the Big Arrow constitute an *antecedent event*. Those on the right constitute a *consequent event*. Nodes are atomic primitives equivalent to concepts. The first node in a kernel indicates its "type" and the other nodes represent parameters associated with that type. Nodes are visualized as "nests" of two-way pointers linking schemas to each other, presumably one nest per concept. Witkowski (1997, pp. 51—53) describes modifications to JCM made in Mott's 1981 ALP system, including adding *motivational kernels* as a kind of intrinsic motivation for influencing agent goals by indicating conditions agents should seek (<HIGH>S) and avoid (<LOW>S)[40].

James Albus leveraged control theory, biology, neural networks, and more[41] to engineer hierarchical systems capable of learning from and reacting to the environment (Albus 2007). His theory of cerebellar function (Albus 1971) proposes how feedback loops between different types of cells in the cerebellum may coordinate motor commands with body positions. Albus et al. (1980) specifies a hierarchical control system consisting of functionally similar, interconnected modules each containing elements for sensory processing, predictive memory, and task decomposition capable of coordinating sensations and actions from raw sensory inputs through increasing levels of abstract behavior. A hierarchical factory control system is described with levels mapping to *output signals*, *action primitives* (e.g., velocity, position, force, torque), *elemental moves* (e.g., reach, grasp, move, release, insert, twist, lock, pull), *simple tasks* (e.g., fetch, mate, fasten), and finally *complex tasks* (e.g., assemble). A detailed outline for a theory of intelligence describing structural and process components (including learning mechanisms of *repetition*, *reinforcement*, and *specific error correction*) is presented in (Albus 1991). Subsequent *multi-resolutional architectures* consisting of similar hierarchies of repeating, interconnected, looping modules include a *real-time control system* (RCS) (Albus 1999), a *4-dimensional real-time control system* (4D/RCS) (Albus and Barbera 2006), and a model of the human brain where "each cortical hypercolumn together with its underlying thalamic nuclei performs as a *Cortical Computational Unit* (CCU) consisting of a frame-like data structure (containing attributes, state, and pointers) plus the computational processes and mechanisms required to build and maintain it" (Albus 2008). The 4D/RCS was "designed to enable any level of intelligent behavior, up to and including human levels of performance in driving vehicles and coordinating tactical behaviors between autonomous air, ground, and amphibious vehicle systems" (Albus 2007). "At the lower echelons, the nodes generate goal-seeking reactive behavior. At higher echelons, they enable goal-defining deliberative behavior" (Albus and Barbera 2006).

[38] Arbib (2018) cites "The Law," "Presbyterianism," or "The English Language" as examples of social schemas. These might also be characterized as *socially codified concepts*.
[39] Becker (1973) did not associate his schemas with Piaget.
[40] See also (Bond and Mott 1981) describing this work applied to the Mark IV Experimental Robot, a.k.a. "Mr Cube."
[41] "Our proposed reference model architecture accommodates concepts from artificial intelligence, control theory, image understanding, signal processing, and decision theory" (Albus and Meystel 2001).

William Powers' (1973) Perceptual Control Theory (PCT)[42] offers a constructivist AI framework based on control theory and the hypothesis that *the purpose of behavior is to control perceptions*. It posits a hierarchy of feedback loops similar to Piaget (1952), Cunningham (1972), and Albus (1980)[43] (and to back-propagation artificial neural networks and to other applications of control theory). He suggests "The entire hierarchy is organized around a single concept: control by means of adjusting reference signals for lower-order systems" (Powers 1973, p. 78). PCT treats agents as being completely self-contained control systems. Hierarchy levels span from *intensity* to *sensation*, *configuration*, *transition*, *sequence*, *relationships*, *program control*, *principles*, and *system concepts* (Vaniver 2015; Forssell 2016, p. 341).[44] Loops at the intensity level process input-output (IO) from sensors. Loops at the sensation level aggregate IOs from the intensity level. The hierarchy continues up through levels of increasing complexity and abstraction.

Ernst von Glasersfeld promoted *radical constructivism*—the principle that agents construct their reality (external world and selves) entirely through the interplay of sense data and motor signals (von Glasersfelds 1974). Von Glasersfeld (1984) describes radical constructivism as "a theory of knowledge in which knowledge does not reflect an 'objective' ontological reality, but exclusively an ordering and organization of a world constituted by our experience." Glasersfeld's main contributions to AI constructivism may be his interpretations and advocacy of the work of Piaget (von Glasersfeld 1974; von Glasersfeld 1982), Ceccato (von Glasersfeld 2001), Powers (Richards and von Glasersfeld 1979), and others. He was particularly aligned with Ceccato (1962), engaging in work on linguistics and machine translation that included the *Multistore System* (von Glasersfeld and Pisani 1970; von Glasersfeld 2001, p. 6). Multistore operationalized Ceccato's ideas by providing rapid matching of elements in service of correlational concept structures. Von Glasersfeld (1984, p. 12) likens *assimilation* to judging the sameness of objects and experiences due to shared attributes and he suggests *accommodation* occurs when attributes are identified that distinguish objects and experiences from one another.[45] He further notes the starting point for perception is when things are isolated as "bounded, unitary objects in the total field of … experience."

Klahr and Wallace (1976) proposed a production system[46] for implementing an *information-processing view of cognitive development* inspired by Piaget and psychological experiments. The goal was to "formulate precise models of performance of the organism at two different levels of development, and then to formulate a mechanism for the transition or developmental mechanisms" (p. ix). They focused on the domain of *quantitative comparison* (QC)—building a series of models that progressively added and adapted production rules consistent with Piagetian staged development. Constituent concepts included class inclusion (CI), conservation of quantity (CON), and transitivity of quantity (TRAN), which emerge from basic productions (operators) of subitizing (Qs), counting (Qc), and estimating (Qe). In the complete model, productions and production systems representing values, attributes, objects, relations, and procedures accumulate in long-term memory and are processed in short-term memories when activated by sensory input or other productions. Schemas develop in order of consistent sequences, common sequences, individual rules, individual productions, production subsystems (*operators*), and production systems[47]. They are organized in a hierarchy of 3 tiers each containing multiple levels, which determine the search order of the productions. Higher tiered/leveled productions, which are more specific, are searched before lower ones, which are more general. *Innate productions* exist at every tier and provide the basic mechanisms for concept construction. They include productions for detecting consistent sequences, transforming common sequences into rules, doing low-level visual and audio encoding, and processing goals. The authors provide "crude starting point" proposals for detecting consistent sequences and transforming sequences into rules (pp. 204 – 207) but such productions were not implemented. The implemented models relied on hand-coded productions. The BAIRN system described by Wallace et al. (1987) leveraged production system elements into an architecture further suited to constructivist learning. It represented knowledge (schemas) as a network of nodes in long-term memory, each representing a feature of the world using several production rules (i.e., *definition list*) and information about the node's connectivity within the network (p. 365, Fig. 8.1). Nodes are highly interconnected such that, "each element in the condition and action of a production is semantically defined at a node elsewhere in the network" (p. 363). Short-term memory structures are used for sensoriperceptual buffering and semantic processing. Adaptation occurs via processes of *node creation*, *node combination*, *redundancy elimination*, and *node modification*. Other innovations included "a limited amount of distributed parallel processing and an explicit treatment of consciousness, motivation and emotion" (p. 359)[48]. As in their previous work (Klahr and Wallace 1976), the QC domain was used to develop and test the system. Other ideas about using production systems for modeling (or implementing) constructivist systems, with closer ties to Piaget, are described by Richard Young (1973, 1974, 1976) and Margaret Boden (1978).

[42] Not to be confused with Kelly's Personal Construct Theory.
[43] Albus and Powers do not appear to have collaborated. However, both published articles about their work in the June 1979 issue of *Byte* magazine where Albus (1979, p. 10) wrote "The brain is first and foremost a control system" and Powers (1979, p. 132): "The key concept behind this revolution [in understanding the nature of all living systems] is control theory."
[44] Forssell (2016) provides a comprehensive summary of PCT.
[45] Von Glasersfeld (1984) uses the terms *elements*, *properties*, and *components* rather than attributes.
[46] Production systems contain rules of the form Conditions -> Actions where Conditions are percepts that trigger the associated Actions.

Rodney Brooks' (1986) *subsumption architecture* may be characterized as having constructivist elements. Schemas correspond to Brooks' finite state machines (implemented as LISP modules) that exchange short messages and are connected in a hierarchy of *levels of competence*. Adaptation is supported by *suppression and inhibition signals* that allow modules at different levels to override other modules' inputs and outputs. Brooks' approach has a nativist character with modules custom-engineered for specific functions[49]. Eight levels of competence suggest staged development—with the lower levels "built" prior to, and supporting, the upper ones. Brooks' (1986, p. 16) levels are:

0) Avoid contact with objects (whether the objects move or are stationary).

1) Wander aimlessly around without hitting things.

2) "Explore" the world by seeing places in the distance that look reachable and heading for them.

3) Build a map of the environment and plan routes from one place to another.

4) Notice changes in the "static" environment.

5) Reason about the world in terms of identifiable objects and perform tasks related to certain objects.

[47] The authors do not use the term *schema*. They state, "The process-structure distinction made by Piaget does not figure in our developmental theory" and "Piaget's structures are replaced by production systems" (p. 189). Also, they rely on the term *equilibration* to characterize Piaget-like adaptation and do not refer to *accommodation* or *assimilation* in a Piagetian sense.
[48] The ACT-R (Anderson and Kline 1977) and SOAR (Laird et al. 1986) cognitive architectures, which, like BAIRN, originated at Carnegie Mellon University, are also production systems based, but much less constructivist given they were designed top-down to cognitively model adults. In the Preface to (Laird et al., 1986), Newell notes SOAR had roots in the Instructible Production System (IPS), which had a goal of being a production system that was "grown, not programmed." General learning mechanisms of *chunking* and *weak methods/subgoaling* were added later to facilitate constructive learning.
[49] The architecture might be extendible to allow modules to be constructively learned, but such methods are not specified or obvious from the paper.

6) Formulate and execute plans that involve changing the state of the world in some desirable way.

7) Reason about the behavior of objects in the world and modify plans accordingly.

Wei-Min Shen (1989, 1994) articulates constructivist AI principles in his work on *autonomous learning from the environment*. Shen (1989) defines learning as "the process of inferring the laws of the environment that allow the learner to solve problems." Shen (1994) states, "Intelligent behavior of any creature, animals or machines alike, is ultimately rooted in its physical abilities to perceive and act in its environment … Every concept or idea of [the] system must eventually have meaning in terms of [its] actions and percepts." He formalizes such systems as consisting of *model applicator* and *model abstractor* algorithms under control of an *integration loop*. The applicator selects actions based on agent goals and predicts new environmental states based on the system's internal (mental) models. The abstractor constructs and revises the models as needed. More generally, models are expressed as a six-tuple $(A,P,S,\varphi,\theta,t)$ where "A is the set of basic actions, P is the set of percepts, S is a set of model states (the internal representation of experience), φ is a state transition function $S \times A \rightarrow S$ that maps a state and an action to the next state, θ is an appearance function that maps states to observations $S \rightarrow 2P$ (2P denotes the power set of P), and t is the current model state of M." Models are built using *m-constructors* (e.g., =, ∧, ∃, +, *). Percepts can be numeric values, objects, features, functions, or relations. Environments are viewed as black boxes represented by triples (Σ, ρ, Δ), where "Σ is a set of inputs, Δ is a set of outputs, and ρ is the environmental mapping function that governs the mapping from the current input to the output." Environments have their own internal logic and can be manipulated by multiple agents.

After discussing various types of environments, models, and learning techniques, Shen (1994) describes the LIVE system that implements the algorithm:

```
Repeat
  Generate a new goal or a new experiment (based on the current model);
  While the goal or experiment is not accomplished:
     Generate a prediction sequence for achieving the goals or experiment,
     Execute the actions in the prediction sequence,
     Perceive information from the environment,
     If there is a prediction failure,
     then find the difference between a success and the failure,
        If some difference is found,
        then call CDL to improve the current model,
        else call CDL+1-like algorithms to create new features or variables.
```

The CDL (*complementary discrimination learning*) algorithm learns by modifying previously learned models when actions yield perceptions that do not match model predictions. Shen calls CDL a *predict-surprise-identify-revise* procedure—anticipating "surprise" as a trigger for making model adjustment in subsequent work such as (Barto et al. 2004; Ranasinghe and Shen 2008; Schmidhuber 2010; Faraji et al. 2016). CDL is shown to learn Boolean concepts, decision lists, prediction rules, and finite automata. The main data structures in LIVE are *prediction rules* that map percepts and actions to changes in the environment. Each rule includes a summary of environmental conditions for a particular state (i.e., percepts), an action the system can take, the predicted change in the environment resulting from the action, and a sibling rule[50]. The rules are characterized by Shen as "c-a-p *production rules* with three components: *conditions, actions, and prediction*." CDL learns conjunctive and disjunctive concepts in an incremental manner via *complementary discrimination*, which performs rule generalization and specialization concurrently using rules and their logical complements. The CDL+1 component can create new features

[50] A sibling rule is a rule that shares actions with the current rule but makes different predictions.

(i.e., rule terms) "that are beyond the scope of the initial perception description language."[51] LIVE environments explored in (Shen 1989) include a hidden-rule Tower of Hanoi environment, child development Balance Beam experiments, and Mendel's pea-hybridization experiments where the system showed "some encouraging results" in each case.

Building on their previous work in psychology and neurobiology (Quartz 1993; Quartz and Sejnowski 1994), Quartz and Sejnowski (1997) summarize evidence for *neural constructivism*, which enables *constructive learning* to occur in brains whereby "the representational properties of cortex are built [progressively] by the nature of the problem domain confronting it." This allows learning to occur in animals through interactions with external environments while not overly relying on innate, specialized, evolutionarily programmed circuits as suggested by nativist and selectionist theories. Their main arguments focus on brain structure changes that occur during cognitive skill acquisition, specifically: changes in synaptic numbers, axonal arborization, and dendritic arborization. Other evidence cited for neural constructivism includes: extensive and protracted postnatal cortical development (occurring well beyond the first two years of life in humans), environmental effects on ocular dominance column development, learning-theoretic models supporting effective incremental learning over time (e.g., Leslie Valiant's PAC), the metabolic efficiency of generating brain structure "as needed" versus retracting unnecessary structure, and neural plasticity in adults (p. 581). Their descriptions of neural constructivism share many characteristics with machine learning practice such as *features*, *feature spaces*, *clustering*, *correlated activity*, *sampling mechanisms*, and *hierarchical representations*—affording potential insights into AI structures and algorithms. Algorithmically, Quartz and Sejnowski (1997, p. 553) suggest: "The general strategy of constructivist learning is this. Rather than start with a large network as a guess about the class of target concepts, avoid the difficulties associated with overparameterized networks by starting with a small network. The learning algorithm then adds appropriate structure according to some performance criterion and where it is required until a desired error rate is achieved."[52] Further discussion about neural constructivism can be found in (Westermann et al. 2007).

Mark Ring (1994) introduces and explores *continual learning*, which focuses on constructivist principles of hierarchical development, unlimited behavior duration, intelligent behavior acquisition, incremental learning, and autonomous behavior. Temporal Transition Hierarchies (TTH) are described which are two-layer neural networks that automatically learn a probabilistic hierarchy of events (sensation-action sequences) based on actions taken by an autonomous agent exploring an environment. Primitive units representing atomic sensations and actions can be combined into sequences of sensations and actions to form higher-level units that are incrementally added to the network. Connection weights between units are adjusted to account for different contexts experienced by the agent. The system is evaluated using environments consisting of simple mazes, the Reber grammar, and the Mozer "gap" task. An agent capable of Continual, Hierarchical, Incremental Learning and Development (CHILD) is introduced that combines TTH with Q-learning. In (Ring 2011), Recurrent Transition Hierarchies (RTHs) are introduced as an improvement over TTH that add recurrent connections to the network to allow an agent to continually learn arbitrary temporal contingencies. Schaul and Ring (2013), show General Value Functions (GVFs) (aka "forecasts"), which are extensions of Sutton's options framework (Sutton et al. 1999), are particularly effective constructivist learning algorithms, superior to Predictive State Representations (PSRs), Temporal Difference (TD) Networks, and TTH with regard to generalization and other properties desirable for continual learning.

[51] Shen points out that new features *must* be present when existing features cannot discriminate between distinct outcomes (e.g., when different classifications are made despite all existing feature values being the same). Such creation of new features is sometimes called *constructive induction*, *automatic feature discovery/engineering/generation* or *discovering hidden features*—essentially what artificial neural networks do by default.

[52] They cite the following "impressive work" on constructivist learners: Azimi-Sadjadi et al. (1993), Fahlman & Lebiere (1990), Frean (1990), Hirose et al. (1991), Kadirkamanathan & Niranjan (1993), Platt (1991), Shin & Ghosh (1995), Shultz et al. (1994), and Wynne-Jones (1993). Subject methods include Cascade-Correlation (CasCor), the upstart algorithm, RAN/GaRBF networks, Ridge Polynomial Networks, and node splitting.

Foner and Maes (1994) extend Drescher (1991) to make schema formation more efficient using *focus of attention* mechanisms. One mechanism, *perceptual selectivity*, "restricts the set of sensor data the agent attends to at a particular instant." Another, *cognitive selectivity*, "restricts the set of internal structures that is updated at a particular instant." Perceptual selectivity relies on spatial and temporal coherence to prune out non-causal inputs. Cognitive selectivity relies on context and goals to prune inapplicable schemas ("facts").

Work by Cangelosi, Schlesinger, and colleagues from their developmental robotics perspective includes Schlesinger (1994) on neural constructivism; Cangelosi and Parisi (1998), Cangelosi et al. (2007), Cangelosi et al. (2010), Morse and Cangelosi (2017) on developmental and agent-based language acquisition; Schlesinger et al. (2000) on constraint-based development of infant motor skills[53]; Schlesinger and Parisi (2001) on online (real-time) sampling in agents that explore their environments; Morse et al. (2010b) on an *Epigentic Robotics Architecture* (ERA) where schemas ("basic ERA units") consist of Self-Organizing Maps (SOM) representing particular features of the environment (e.g., color, shape, body posture, word representations) connected to "hub" SOMs weighted using positive Hebbian learning[54]; Cangelosi et al. (2000), Coventry et al. (2005), Cangelosi and Riga (2006), Marocco et al. (2010), Cangelosi (2010), Stramandinoli et al. (2012), Stramandinoli et al. (2017) on sensorimotor grounding of words and concepts; and Schlesinger (2020) on computational models of development classified as connectionist, dynamic field theory-based, rule-based, and Bayesian. In addition to SOMs per (Morse et al. 2010b), they used Recurrent Neural Networks (esp. Jordan Networks) to integrate linguistic, visual, and proprioceptive inputs for grounded concept learning (Stramandinoli et al. 2012; Stramandinoli et al. 2017). Cangelosi and Schlesinger's (2015) book on developmental robotics is a rich source of information about constructivist AI that describes work in developmental psychology that can inform agent development. Six "experiment-focused" chapters look at relevant psychological models and experiments grouped into topics of: novelty, curiosity, and surprise; perceptual development; motor development; social learning; language; and abstract knowledge. Appendix A herein summarizes topics, keywords, and references covered in that book.

Sandra Marshall (1995), inspired by the psychology of Bartlett (1932) and of Piaget, and the AI formalisms of Rumelhart, Minsky, and Schank (i.e., schemas, frames, and scripts), focuses on schemas, which she defines as:

> *A vehicle of memory, allowing organization of an individual's similar experiences in such a way that the individual*
>
> - *can easily recognize additional experiences that are also similar, discriminating between these and ones that are dissimilar;*
> - *can access a generic framework that contains the essential elements of all of these similar experiences, including verbal and nonverbal components;*
> - *can draw inferences, make estimates, create goals, and develop plans using the framework; and*
> - *can utilize skills, procedures, or rules as needed when faced with a problem for which this particular framework is relevant.* (Marshall SP 1995, p. 39)

Accordingly, each schema contains four types of knowledge: identification knowledge, elaboration knowledge, planning knowledge, and execution knowledge. Marshall SP (1995, pp. 377 - 390) presents a hybrid model where a schema consists of one connectionist network and three production systems[55].

[53] This work explores the idea that assimilation and adaptation of motion schemas (i.e., "movement primitives") can progressively build a repertoire of sophisticated motor abilities. Degrees of freedom of motion become enabled gradually (or are "frozen") to limit the search space and develop schemas suitable for chaining.

[54] They also discuss using Echo State Networks as dynamic reservoirs in ERA units to better address temporal and nonlinear relationships.

[55] It is instructive to note similarities between Marshall SP (1995) and Klahr and Wallace (1976). Both describe hybrid models that use production system and connectionist elements. Both were strongly influenced by Piaget and developmental psychology literature and focused on arithmetic concept domains.

Identification knowledge uses a neural net classifier (the connectionist network)[56] to recognize the type of experience (problem or situation) at hand. Once recognized, relevant input from the environment is parsed into *clause encodings* containing information—like owner, object, and time—appropriate for executing rules contained in the production systems. The four components are connected via a blackboard where inputs, outputs, and queries are shared to allow situations to be parsed, evaluated, and action taken using knowledge specific to each schema. The identification network can iterate on current situation data to refine the assessment of the best matching situation, thus determining which productions are activated. Beyond suggesting that identification networks can be learned using backward propagation, Marshall does not address schema learning (adaptation). Production system components were manually programmed to handle the elaboration, planning, and execution appropriate for each schema. Marshall alludes to staged development in describing a hierarchy of levels progressing from *microfeatures*, to *single knowledge components*, to a *full schema*, to *collections of schemas* (p. 392).

Mark Witkowski, working with Bond and Mott (1981), leveraged Becker's (1973) constructive approach in developing the Mark IV Robot. Witkowski (1997) proposes a *Dynamic Expectancy Model* (DEM) that provides "a novel form of learning by reinforcement," which uses *expectancies* (predictions, *micro-* or *μ-hypotheses*) to control agent learning via *μ-experiments*. Influenced by animal behavior literature, the DEM accounts for both innate and learned capabilities. Witkowski (1997), particularly inspired by Tolman (1932, 1948) and MacCorquodale and Meehl (1953), builds on the implementations of Becker (1973), Bond and Mott (1981), and Drescher (1991). DEM processing is described as "a low level version of a scientific discovery process" (Witkowski 1997, p. 58)—reminiscent of Kelly (1955) and Shen (1984). μ-hypotheses are the basic units of learning (i.e., schemas). They are reinforced or extinguished over time based on the success or failure of their predictions. Sensory inputs are processed into *tokens*, which are incorporated into constructs called *signs* and *sign sets*. Internal *symbols* can cause the output of *actions* and *compound actions*. Adaptation processes include *creation*, *corroboration*, *reinforcement*, *differentiation*, and *forgetting* of μ-hypotheses. The model was implemented using the *SRS/E algorithm*, which generated *Dynamic Policy Maps* (DPMs) and operated similarly to Sutton's (1990) Dyna-Q. SRS/E stands for "Stimulus-Response-Stimulus/Expectancy" to reflect the key elements of each μ-hypothesis: the current state or context (Sign1) of the agent, an action (Response1) the agent can take, and a consequent state (Sign2) resulting from the action. Schemas are encoded using seven interrelated lists comprising: input tokens, signs denoting environmental states, responses, behaviors, goals, hypotheses, and predictions. The hypothesis list is used to construct DPMs. The lists are initialized to reflect the agent's innate capabilities (i.e., the *ethogram*) (Witkowski 1997, p. 57) and are continually adjusted thereafter to account for learning and other changes to the agent. SRS/E was shown to perform well on latent learning and place learning tasks. Witkowski (2007) proposes an *action-selection calculus* specifying three ways states can be connected to actions; five rules for how actions, signs, and predictions are updated; and four rules for how learning occurs. It claims to unify the five classic learning theories: stimulus-response behaviorism, associationism, classical conditioning, operant conditioning, and Tolman's sign-learning model.

Daniel Wolpert and colleagues focused on biological motor control and motor learning. Wolpert et al. (1995) present a *Kalman filter model* consistent with cerebellar function that combines a *forward model* that predicts motor states with a *sensory output model* that predicts sensory feedback and together reliably estimate the effects of motor commands. Wolpert et al. (1998) examine various cerebellar learning models and conclude, "The cerebellum contains *multiple pairs of corresponding forward and inverse models*[57], each instantiated within a microzone[58]". The modules (schemas) also include a *responsibility*

[56] The neural net was a 3-layer feedforward network with 27 input nodes, 14 hidden units, and 5 output nodes. Each problem was coded as 27 binary features used to classify 5 arithmetic problem types: Change, Group, Compare, Restate, Vary. The clause encodings and rules for the production systems were engineered for this domain.

[57] Forward models "mimic the causal flow of a process by predicting its next state (for example, position and velocity) given the current state and the motor command." Inverse models "invert the causal flow by estimating the motor command that caused a particular state transition." (Wolpert et al. 1995)

[58] Microzones are attributed to be the basic functional units of the cerebellum (Oscarsson 1979; De Zeeuw 2021).

*detector* that works in concert with the forward model to coordinate appropriate activation of multiple modules for effecting composite behaviors. Wolpert et al. (2011) summarize components, processes, and representations suitable for motor learning. Components determine how information extraction from the environment occurs (e.g., what elements are fixated on versus filtered out, how processing delays and noise are accounted for), what decisions and strategies are appropriate for executing given tasks (i.e., resolving interactions between sensorimotor, perceptual, and cognitive components), and how different classes of control contribute to generating appropriate motion sequences (i.e., predictive control, reactive control, and biomechanical control). Processes enabling motor learning include error-based learning (i.e., dynamically correcting for mismatches between goal states and actual states), reinforcement learning (e.g., using reward signals to drive actions), use-dependent learning (i.e., habituation), and observational learning (i.e., learning by observing others). Representations suitable for motor learning are characterized as being mechanistic or normative models. The former tend to be built from *motor primitives* (primitive schemas), which combine together using *generalization functions*. The latter tend to rely on *credit assignment* mechanisms (e.g., Bayesian analysis) to calculate how multiple underlying causes determine actions. McNamee and Wolpert (2019) provide a unifying account on how the brain may employ internal models for motor control using *Bayesian inference* and *optimal feedback control*.

Yiannis Demiris leveraged Wolpert and Kawato (1998), Meltzoff and Moore (1989, 1997), and others to develop robots capable of learning motor behaviors using imitation and social perception. Demiris and Hayes (1996) present a biologically inspired architecture for *imitative learning* consisting of modules for visual preprocessing, proprioceptive analysis, relationship establishment, movement analysis, and movement matching. Per the architecture, an *imitator* visually collects posture data from a *demonstrator* and proprioceptively reproduces the motions it perceives in itself. Demiris and Hayes (2002) present a modular, *dual-route* architecture for motion learning that combines *active* and *passive* imitation. Each module, operating in parallel, pairs a previously learned behavior (inverse model) with a forward model to attempt to match demonstrated behaviors. The forward models simulate (imagine) the demonstrated behavior, predict the next observed states, and generate an error signal to indicate how well the actual next states matches the prediction. Errors accrue as a *confidence factor* for each behavior, which enables the imitator to determine the best-matching behavior. If no demonstrated behavior is matched with high confidence, a new schema is generated via passive imitation as described in (Demiris and Hayes 1996). The architecture was tested using simulated robots performing arm movements conforming to the international standard semaphore code.[59] Demiris and Khadhouri's (2006) HAMMER architecture arranges the modules (schemas) from (Demiris and Hayes 2002) into distributed hierarchies to enable complex and abstract behaviors to arise from more primitive behaviors. It relies on a *top-down control of attention mechanism* to determine what state information to collect from the environment that is appropriate for each schema. The forward model components determine what information is required to run the forward model simulations and that data is then used to drive the inverse model components. Experiments were conducted where an imitator perceived an action performed by a demonstrator and executed the best matching behavior from a set of eight previously programmed behaviors. Behaviors (forward models), represented as Baysian networks, can be learned using motor babbling as described in (Dearden and Demiris 2005).

Joshua Tenenbaum (1999) proposes a Bayesian framework for concept learning that supports learning from few or many examples, thus uniting similarity- and rule-oriented generalization[60]. It represents early work using Bayesian methods to understand human intelligence and build intelligent machines. Kemp and Tenenbaum (2008) describe a Bayesian model that discovers forms and structures for representing environmental data. They promote the view that different knowledge representations (schemas/models) are needed for different types of knowledge. Learnable forms include partitions, chains, orders, rings, hierarchies, trees, grids, and cylinders. Tenenbaum et al. (2011) discuss how "abstract knowledge encoded in a probabilistic generative model" can serve to constrain a hierarchy of models to more quickly achieve learning and induction. The authors suggest *Hierarchical Bayesian models* may be able to learn *framework theories* in key domains like intuitive physics, psychology, and biology, and lead to an answer to the (constructivist) question: "How can domain-general mechanisms of learning and representation build domain-specific systems of knowledge?" Ullman and Tenenbaum (2020) argue for harmonizing cognitive development and computational modeling to answer key questions underpinning constructivist AI. They discuss how hierarchical Bayesian models can account for learning in infancy and childhood by leveraging core knowledge about objects, agents, space, and time in infancy and building intuitive theories during childhood. They suggest implementing knowledge and theories as *probabilistic generative programs*. Like Kelly (1995), Shen (1989), and Witkowski (1997), they liken learning to *naïve science*: a (Bayesian) process where an agent hypothesizes representations (models) of the environment that are then assessed, refined—and maybe discarded—over time. They introduce a "child as hacker" metaphor likening learning agents to software developers, discussed in (Rule et al. 2020). Although Tenenbaum might balk at being called a constructivist[61], his approach is aligned with AI constructivism and draws heavily on cognitive and developmental psychology research.

[59] See https://en.wikipedia.org/wiki/Flag_semaphore.
[60] Tenenbaum defines the rule-oriented approach as "hypothesis testing in a constrained space of possible rules" and the similarity-oriented approach as "computing similarity to the observed examples" (Tenenbaum 1999, p. 2).

Juyang Weng et al. (2001) promoted AI constructivism as a "new field" called *Autonomous Cognitive Development* citing Weng et al.'s (1999) robotic Self-organizing, Autonomous, Incremental Learner (SAIL) as a prototype embodying key constructivist principles. SAIL was engineered to learn representations and architectures through self-exploration and human teaching, enabling the robot to autonomously navigate unknown, unconstrained environments and to recognize and reach for objects. Motor movements were learned primarily by a trainer physically manipulating the robot to associate sensory input states with appropriate motor outputs. External reinforcement by human trainers pressing "good" or "bad" buttons and intrinsic reinforcement provided through a *value system* were also used (Huang and Weng 2002). Agent state consisted of sensor readings recursively convolved with resampled versions of themselves to account for different temporal contexts. These states were mapped to motor values using an Incremental Hierarchical Discriminating Regression (IHDR) tree algorithm (Weng et al. 1999; Weng et al. 2001; Weng and Hwang 2007). IHDR is claimed to unify classification and regression[62] and (1) Handle high dimensional inputs, (2) Perform one-shot learning, (3) Dynamically adapt to increasing complexity, (4) Avoid getting stuck in local minima, (5) Operate incrementally, (6) Effectively maintain long term memory, and (7) Is suitable for real-time operation (W&H 2007, p. 3). IHDR maintains clusters of inputs (x-clusters) linked to associated outputs (y-clusters) at nodes of an oblique decision tree.[63] For each input-output sample ($x_i$, $y_i$), $y_i$ is used to find the closest y-cluster via Euclidean distance. The identified y-cluster indicates which x-cluster sample ($x_i$, $y_i$) belongs to. $x_i$ and $y_i$ are then used to update the statistics of their corresponding clusters. (If a $y_i$ value is not provided, the $x_i$ vector is used to find the "best" y to output from a tree leaf.) Child nodes are spawned (or searched) when finer granularity in input-output mapping is required as indicated by the statistical match of the current sample to each x-cluster. *Discriminating feature subspaces* are calculated using the centers of x-clusters to determine which child nodes should be searched—thus focusing on the most discriminating features in $x_i$ for the current sample. IHDR is claimed to model the brain's associative cortex (i.e., the area between the primary sensory cortex and motor cortex) wherein *layers* represent different cortical layers, *nodes* represent cortical patches, and *clusters* represent neurons (W&H 2007, p. 24).

---

[61] References to Piaget are mostly absent in his work. Tenenbaum et al. (2011) characterize "constructivism" (and "theory theory") as "less formal approaches to describing the growing minds of children." For an analysis of synergies between probabilistic models and Piaget's work see (Tourman, 2016). Learning as a form rational Bayesian inference has also been explored as *rational constructivism*. See (Xu F et al., 2012).

[62] Weng and Hwang (2007, p. 4) define regression as a task "similar to the corresponding classification one, except that the class label $l_i$ is replaced by a *vector* $y_i$ in the output space." This differs from more typical definitions of regression where a vector input yields a *scalar output*. (See https://en.wikipedia.org/wiki/Regression_analysis.) The composition of "y-clusters" is unclear. In their main use case of robot motor control, Weng appears to consider an output vector to be a temporal sequence of values of one output variable—not a vector of different output variables. Multi-output regression (Borchani et al. 2015) and multi-output learning (Xu D et al. 2019) are more typical of what is meant by regressing to vector outputs.

[63] Oblique trees use multiple input features for node splits.

IHDR trees may be construed as schemas, with node levels corresponding to stages of development and x- and y-cluster constructs within nodes providing fine-grained structure. The learning (adaptation) algorithms are provided by eight procedures detailed in (W&H 2007). Schemas (trees) are further organized into levels where the lowest level encodes innate behaviors (e.g., visual motion detection and tracking). Such schemas can be learned offline using IHDR operating as a "prenatal learning process." Higher-level schemas incorporate progressively more temporal context and are learned (online) through interaction with the environment. Learned behaviors have priority over innate behaviors and are overridden by innate behaviors if an adequate learned behavior does not exist. (In general, the behavior having the highest confidence of being correct is executed.) In a more comprehensive architecture, Weng and Hwang (2006, Fig 4) describe stackable modules consisting of a sensor-to-effector mapper, stored context prototypes, a value controller, an attention selector, a motor mapper, and a delay mapper.

Later work by Weng et al. includes Weng and Luciw (2009) on Lobe Component Analysis (LCA)—a cortex-inspired theory of Hebbian learning claimed to be spatially and temporally optimal; Ji et al. (2008) and Weng and Luciw (2014) on visually extracting concepts like object location and type from cluttered scenes using *concept networks* (a.k.a. *developmental networks* and *where-what networks*) that process information bottom-up, top-down, and laterally; and Weng (2018) and Wu X et al. (2021) that characterize schemas as specialized Turing Machines learned by developmental networks.

Lee McCauley (2002) presents a *neural schema mechanism* to transform Drescher's (1991) constructivist system into more of a connectionist network. The mechanism defines *item nodes* to represent states of the environment, *action nodes* to represent actions that can be taken by the agent, *schema nodes* to specify relationships between items and actions, and *goal nodes* to control node activations. Nodes are connected using *context links*, *result links*, *action links*, *goal links*, *host links*, and *none links*. Unlike typical neural networks, node activations depend on the node and link types, e.g., link weights are determined by values of *relevance*, *reliability*, *correlation*, and *desirability*. Schema nodes maintain statistics for determining which actions to select and control production of new nodes (i.e., spin-off schemas, accommodation). An agent is said to be "conscious" of nodes that exceed a certain level of activation. "Conscious broadcasts" cause nodes related to those currently in the "spotlight" to increase activation potentials, which increase chances they cause an action to be selected so "an agent can discover new paths to a solution or goal state." McCauley implements and evaluates the system using Russell and Norvig's (1995) Wumpus World.

Jeff Hawkins (2004) does not focus on childhood development nor cite Piaget but subscribes to constructivist principles in his efforts to build intelligent machines:

- "The senses create patterns that are sent to the cortex, and processed by the same cortical algorithm to create a model of the world (p. 44)."
- "When you are born, your cortex essentially doesn't know anything. … All this information, the structure of the world, has to be learned (p. 111)."
- "[We] have to train the memory system much as we teach children. Over repetitive training sessions, our intelligent machine will build a model of *its* world as seen through *its* senses (p. 141)."

Persistent themes in Hawkins' work include a focus on the human neocortex as the seat of intelligence (Hawkins 2004), sequence learning and prediction as keys to cognition (Hawkins 2004; Hawkins and Ahmad 2016), cortical columns running a common algorithm as the essential processing units (Hawkins 2004; Hawkins et al. 2017b), use of sparse distributed representations for information coding and processing (Hawkins 2004; Ahmad and Scheinkman 2019; Numenta 2021a), and utilizing world models and reference frames for constructing intelligent agents (e.g., grid cells and allocentric representations) (Lewis et al. 2019; Klukas et al. 2020; Hawkins 2021; Lewis 2021).

Hawkins' (2004) *memory-prediction framework* proposes a hierarchical auto-associative memory that stores sensorimotor sequences using invariant feature representations. That memory is used to predict what occurs next in newly encountered sequences. Signals flow up and down sensorimotor hierarchies to affect pattern learning, recognition, and behavior[64]. The basic learning functions are identified as *classification formation* and *sensorimotor sequence construction* and are roughly mapped to the mammalian cortex, hippocampus, and thalamus. Hawkins provides examples of human learning and suggests how the model may accommodate them, including visual saccades and fixation, language and music understanding, memory, environment navigation, and sensorimotor control of various forms. In young brains, Hawkins suggests memories are stored higher up in the cortical hierarchy and are thus slower to react than mature brains since "it takes time for the neural signals to travel up and down." He further notes young brains have "not yet formed complex sequences at the top and therefore cannot recognize and play back complex patterns." As the brain gains experience, it re-forms memory representations further down in the hierarchy which "frees up the top for learning more subtle, more complex relationships."

The memory-prediction framework was implemented as the Hierarchical Temporal Memory (HTM) model (Wikipedia, Memory-prediction framework; Hawkins et al. 2017a; Hole and Ahmad 2021). The first generation focused on spatial and temporal pooling of input patterns for learning, and probabilistic pattern matching for inference. Cortical learning algorithms (CLAs) formed the core of second-generation HTM implementations. CLAs more closely modeled layers and mini-columns of the cerebral cortex, accounting for the formation and decay of synapses. Key elements of this generation of HTM included *sparse distributed representations* (SDRs)[65], a spatial pooling algorithm, and a sequence memory algorithm. A later generation of HTMs added a theory of sensorimotor inference that proposed "cortical columns at every level of the hierarchy can learn *complete models of objects* over time and features are learned at specific locations on objects" (Wikipedia, Hierarchical temporal memory). HTMs utilize sensorimotor feedback signals, feed forward signals, and context signals. Hawkins notes HTM networks need not integrate motor control—rather, they can learn to predict changes in the environment from sensory sequences occurring in the environment independently of any actions by the sensing agent. The latest incarnation of the work—the *Thousand Brains Theory*—posits "the brain uses maplike structures to build a model of the world—not just one model, but tens of thousands of models of everything we know" (Numenta 2021b).

Kristinn Thorisson (2012) takes a system-level view of AI constructivism, deprecating what he calls *construc**tion**ism* in favor of constructivism. The former is characterized as "systems whose gross architecture is mostly designed from the top-down and programmed by hand." He claims to take Drescher (1989) a step further by requiring an AGI *architecture* automatically grows from *seeds* and takes into account factors of "temporal grounding, feedback loops, pan-architectural pattern matching, small white-box components, and architecture meta-programming and integration" (Thorisson 2012, pp. 160-161). Key elements of Thorisson's view were implemented in the *Autocatalytic Endogenous Reflective Architecture* (AERA) (Nivel et al. 2013; Nivel et al. 2014a). In AERA, schemas (*models*) consist of left-hand (LH) patterns, right-hand (RH) patterns, and *guard equations*. LH patterns are said to *predict* (also *produce* or *cause*) associated RH patterns. Patterns consist of sets of features (facts, states, concepts) sensed from or acting upon the environment. For example, an LH pattern might be ("A is_bus", "A has_color yellow", "A bears_number 19", "A bears_license_plate_number SX445") and an associated RH pattern might be ("A stops_at city_university")[66]. Guard equations assign values to variables between RH and LH patterns based on the kind of reasoning being done. Deductive reasoning from causes (LH) to predictions (RH) uses forward chaining. Abductive reasoning from goals (RH) to causes (LH) uses backward chaining (Nivel et al. 2013, p. 14). Both occur simultaneously. The LH-RH schema structure is also used to represent discrete facts known by the agent. Some schemas are innate[67] but most are learned and are retained if they make successful predictions (confirmed from acting on the environment). Nivel et al. (2013, p. 24) write, "Model acquisition is triggered by either the unpredicted success of a goal or the failure of a prediction." Piagetian-like assimilation and accommodation (*abstraction*) is supported whereby "a new model is copied from the original model and differences between values held by the conflicting evidences are represented by variables introduced in the new model" (Nivel et al. 2013, p. 20). Sequences of schemas are chained to form more complex schemas and can be "compressed" to form more compact schemas. An *executive* program provides the inference engine that computes predictions and goals, creates and manages models, focuses attention, manages resources, and performs other functions. Promising results were achieved in experiments where an AI called *S1* learned conversational speech and body movements by observing humans (Nivel et al. 2013, pp. 35-51). The initial AERA incarnation relied on learning by observation.

---

[64] Roughly, expectation signals flow down the hierarchy and are compared with sensorimotor signals that flow up.

[65] SDRs are the main knowledge representations in the HTM model. They are large sparse binary vectors where each bit, corresponding to a different neuron, represents a different learned feature. SDRs with similar bit patterns are semantically similar. "HTM theory defines how to create, store, and recall SDRs and sequences of SDRs."

[66] This example was paraphrased from (Nivel 2013, p. 13). In (Nivel et al. 2013, p. 33) the general form of a model (schema) is presented as "($LT(Q0,Q1,...,Qn,T0,T1)$, $RT(P0,P1,...,Pm,T2,T3)$) where $Qi$ and $Pj$ are variables representing arbitrary quantities, $T0$ and $T1$ are variables that define the time interval within which L holds, and $T2$ and $T3$ variables defining the time interval for R." In a 2013 lecture (AGI-13 Summer School – AERA 6), Nivel gives an example of a model where LT = robotic gripper move command with time and control parameters and RT = resulting gripper location with time parameters.

Michael S. P. Miller (2013a, 2013b, 2018) draws directly on Piaget's later work (Piaget 1977, 1985) in specifying his *Piagetian Modeler* (a.k.a. *Piagetian Autonomous Modeler*). Piaget-defined processes of *observation*, *coordination*, *reflection*, and *consolidation* comprise the Piagetian Modeler's main functional units. Each unit (*pattern*) is decomposed into subpatterns—yielding a full-featured cognitive architecture consisting of many diverse specialized modules. Schemas are networks of memory elements called *neural propositions* organized collaterally and hierarchically to encode models of the world. Biologically, Miller claims elements of a neural proposition map to the axon, soma, and dendrites of a neuron. Logically, neural propositions are "cognitive containers" called *schemes* (also *coordinations*) that are *affirmed* or *negated* using *markers* and have 1 to n links to other schemes[68]. Scheme types include *internal*, *external*, and *inference* schemes. Internal and external schemes *represent* (link to) internal and external *observables* at the lowest level of the cognitive hierarchy, i.e., *propriocepts* and *exterocepts* respectively. Inference schemes link to internal and external schemes and to other inference schemes at increasingly higher levels of the hierarchy (also forming sequences). Each inference scheme represents a feature, action, event, situation, episode, concept, hypothesis, goal, or other mental element learned by the agent. Miller operationalizes other terminology from Piaget including *subsystems*[69], *actions*, *circular reactions*, *operations*, *groups*, and *totality*. Schemas are learned through assimilation and accommodation acting upon schemes. Implementation details and experimental results are forthcoming.

Maria Hedblom and collaborators (Hedblom et al. 2015; Hedblom 2018) formalized Lakoff-Johnson image schemas as *families (sets) of interlinked theories* that could be used for constructing new concepts via *conceptual blending* (Fauconnier and Turner 1998) and for facilitating symbol grounding. In (Hedblom et al. 2015) they use the DOL language of Mossakowski et al. (2013)[70] and Common Logic (ISO/IEC 24707) for encoding image schemas—leveraging formal ontologies to represent agent knowledge. They focus on schemas related to *path following* and *containment* and show how low-level schemas like MOVEMENT_OF_OBJECT, MOVEMENT_ALONG_PATH, MOVEMENT_IN_LOOPS, REVOLVING_MOVEMENT, and CLOSED_PATH_MOVEMENT can combine and specialize through the addition of features (*spatial primitives*) like PATH, START_PATH, END_PATH, FOCAL_POINT, and LANDMARK (Hedblom et al. 2015, Fig. 2). Primitive schemas were hand coded, not learned. Concept construction (learning) was investigated as a higher-level blending of schemas. A Common Logic sample for encoding the MOVEMENT_OF_OBJECT schema is:

[67] Innate schemas consist of "a small *seed*, containing … "drives" (i.e., mission goals and constraints) and a relatively small amount of knowledge to bootstrap learning" (Nivel et al. 2014b, p. 2).
[68] This summary relies on introductory material in (Miller 2018) that supersedes (Miller 2013a, 2013b).
[69] Subsystem is synonymous with schema in Piaget and Miller's descriptions.
[70] The Distributed Ontology, Modelling and Specification Language (DOL), aims at providing a unified meta language for handling multiple ontology languages such as OWL, RDF, OBO, Common Logic, and F-logic.

```
(forall (m)
  (iff
    (MovementOfObject m)
    (exists (o)
      (and
        (Movement m)
        (Object o)
        (has_trajector m o)))))
```
[71]

Hedblom et al. (2015) describe how image schemas related to path following can lead to concepts like "stream of consciousness," "train of thought," and "line of reasoning" using conceptual blending. They show how the concept THRILLER can emerge from concepts of STORY and ROLLER_COASTER which both share aspects of the more primitive schema SOURCE_PATH_GOAL. In (Hedblom 2018), a Two-Object schema family is presented that encompasses schemas CONTACT, SUPPORT, and LINK, which inherit from schemas VERTICALITY and ATTRACTION. The Image Schema Logic ($ISL^{FOL}$) language is introduced for representing image schemas, which leverages Region Connection Calculus (RCC)[72], Qualitative Trajectory Calculus, and Linear Temporal Logic. Hedblom et al. (2021) suggest $ISL^{FOL}$ may be used to improve robot performance in unfamiliar environments by facilitating reasoning about functional relations, enabling reasoning about alternatives to a plan, increasing adaptability through analogy, and improving natural language understanding. Examples of $ISL^{FOL}$ for encoding MOVEMENT_OF_OBJECT and MOVEMENT_ALONG_PATH schemas from (Hedblom 2018) are:

```
  ∀O:Object □(MOVEMENT_OF_OBJECT(O) ↔ Move(O))□
  ∀O:Object, ∀P:Path (MOVEMENT_ALONG_PATH(O, P) ↔
Move(O) ∧ CONTACT(O, P) U(¬(Move(O) ∨ CONTACT(O, P))))
```

Aguilar and Perez y Perez (2015, 2017), working in the field of computational creativity, present a computational model called *Developmental Engagement-Reflection (Dev E-R)* that approximates Piaget's assimilation and accommodation processes. In (Aguilar and Perez y Perez 2015), they model visual development for a simple agent in a virtual 3D world. Endowed with monocular color vision and an ability to move its head in 9 front-facing directions, the agent develops the ability to recognize a multitude of color and size concepts from an initial capacity to recognize 3 color luminosities (red, green, blue) and 2 sizes (big and small). Learning occurs through repeated exposures to objects in the environment. In the case of colors, a count of each primitive color feature is incremented each time the feature is observed. The color is recognized as a salient feature when it reaches a threshold count. That feature is then used as the basis for recognizing new colors by distinguishing (and counting) lighter and darker versions observed in the environment. A similar process generates size concepts. Starting with recognizing a distinction between big (B) and small (S) stimuli, and counting such occurrences, finer grain size features emerge based on repeated exposure to new size stimuli (e.g., B2, B3, S2, S3). Object recognition emerges by biasing the agent to attend to stimuli that move, have bright colors, are novel, and are otherwise affective, i.e., elicit pleasure, displeasure, surprise, or cognitive curiosity.

Once created, features are available for agent *contexts* and *schemas*. Contexts are memory structures containing: (1) features of the object at the current center of attention of the agent, (2) affective responses associated with the object, and (3) the agent's current expectations for feature change (if the context is a *current-context*). Schemas are either *basic* or *developed*. Basic schemas encode innate behaviors using *context* and *associated action* elements. Developed schemas encode behaviors learned through environmental interaction. They have the same structure as basic schemas but add an *expected context* element. The engagement-reflection process creates and maintains schemas. *Engagement* attempts to match cur-rent contexts with contexts in stored schemas. If a suitable match is found and the associated action defined in the schema generates the expected result, the agent is in cognitive equilibrium and the selected schema is considered stable. If a suitable context match is not found, the agent is considered to be in disequilibrium and a *reflection* (accommodation) process occurs using *generalization* or *differentiation* to modify an existing schema or create new ones. Schemas are distinguished by how many specific (instantiated) context features they have. Those with more instantiated features are more concrete. Those with fewer instantiated features are more general. Schemas were generated that allowed an agent to attend to pleasurable objects in its environment. Aguilar and Perez y Perez (2017) extend the system to include tactile capabilities, i.e., hand movements, grasping, and touch sensing of object contact and textures.[73] They report that the agent learned schemas that allowed it to visually follow its own hand movements and coordinate reactions to visual and tactile stimuli.

[71] Note this schema uses OBJECT, MOVEMENT, and HAS_TRAJECTOR primitives.
[72] RCC defines 8 relationships between spatial regions: disconnected (DC), externally connected (EC), equal (EQ), partially overlapping (PO), tangential proper part (TPP), tangential proper part inverse (TPPi), non-tangential proper part (NTPP), and non-tangential proper part inverse (NTPPi).

Google DeepMind research relevant to AI constructivism includes: Higgins I et al. (2016) on using a variational autoencoder (VAE) that learns disentangled features to mimic visual feature learning in humans; Battaglia et al. (2016) on an *interaction network* that decomposes environments into networks of objects and relations and reasons about them explicitly; Higgins I et al. (2017) on the *Symbol-Concept Association Network* (SCAN) that extends the VAE architecture to extract abstract concepts grounded in disentangled features from visual input, associate those concepts with symbols, and enable construction of novel concepts; Hill et al. (2017) on understanding grounded language learning in neural networks that link visual features (i.e., shapes, colors, patterns, shades, sizes) with words in a way that parallels human language acquisition; Rabinowitz et al. (2018) on using meta-learning to build a neural network (*ToMnet*) that learns Theory of Mind models for different species of agents; Rao et al. (2019) on *Continual Unsupervised Representation Learning* (CURL) that learns tasks (concepts) without task or class labels using a mixture-of-Gaussians latent space, dynamic expansion, and mixture regenerative replay; Groth et al. (2021) on *SelMo*—a (self-motivated) system that acquires and retains skills through optimized, curiosity-based, off-policy exploration of its environment; and Jaegle et al. (2021) on *Perceiver IO* which adds a cross-attention mechanism to the Perceiver (Transformers) model to produce a system that scales linearly with multiple inputs (e.g., images, audio, natural language) and multiple outputs (e.g., text prediction, image classification, optical flow fields, audiovisual sequences) and may hold promise as a general purpose neural network architecture. Piloto et al. (2022) present the PLATO model (for Physics Learning through Auto-encoding and Tracking Objects) that learns the intuitive physics concepts of solidity, object persistence, continuity, unchangeableness, and directional inertia from synthetic video data.

Constructivist-oriented work by Vicarious[74] includes:

Kansky et al. (2017) and Vicarious (2017a) describe *Schema Networks*, inspired by Drescher (1991), which learn causal relationships and reusable concepts from sensory data. Schema Networks rely on object instances (*entities*) and associated attributes extracted from videos to predict attribute changes due to environmental actions. Schemas consist of entity-attribute and action variables logically connected to future entity-attribute states[75]. Multiple schemas are combined into a Schema Network capable of probabilistically predicting entity-attribute changes in a Markov Decision Process (MDP) framework. "Breakout" video game variations are used as learning and testing environments. The authors report (model-based) Schema Networks are superior to the model-free Asynchronous Advantage Actor-Critic (A3C) algorithm and Progressive Networks with respect to training efficiency, zero-shot generalization, robustness, and learning transfer.

George et al. (2017) describe generative models for visual object recognition using *Recursive Cortical Networks* (RCNs). In this work, RCNs are used to understand letterforms in a general purpose "common sense" way by modeling objects as a combination of contours and surfaces via a network of features, pooling nodes, and lateral connections. The authors used RCNs to solve text-based CAPTCHAs[76] in 2013. RCNs build on ideas used in other compositional models to create structured probabilistic graphical models amenable to inference using Belief Propagation. Characterized as providing *scaffolding* versus being a *tabula rasa* approach (Vicarious 2017c), the system is engineered to prioritize and distinguish (factorize) contour and surface features. This enables RCNs to be trained using orders of magnitude fewer examples than other neural networks. Unlike convolutional neural network CAPTCHA breakers, training data are clean examples of letters from representative fonts, as opposed to large sets of distorted examples.

[73] Sensed features were {color, size, movement, position} for vision and {texture, hand_open_or_closed} for touch.
[74] https://www.vicarious.com/science/. Vicarious was acquired by Alphabet in 2022 and folded into its Intrinsic subsidiary. Their original goal was to create generally intelligent systems.
[75] Logical connections are comprised of conjunctions, disjunctions, and "self transitions."

Stone et al. (2017) and Vicarious (2017b) propose a method for improving object recognition in computer vision applications using convolutional neural networks (CNNs) by understanding the compositionality of a scene. Understanding compositionality entails the ability to learn and recognize objects and parts of objects in a decomposable, reusable manner. By masking out distinct objects that exist in close spatial proximity[77], they "encourage networks to form representations that disentangle objects from their surroundings and from each other"—a property not provided by current CNNs. They found no negative effects on performance from filtering out context information for the medium- to large-sized objects they tested and claim their method "is a step towards making CNN-based representations more amenable to explicit context modeling through an external mechanism (by cleanly separating the representation of objects from their context)."

Hay et al. (2018) and Vicarious (2018) address learning abstract concepts called Sensorimotor Contingencies (SMCs) through interacting with the environment. Whereas Schema Networks (Kansky et al. 2017) are networks of entities and attributes linked by state transition probabilities, SMCs are small programs that encode perception-action sequences as hierarchies of concepts similar to the options of Sutton et al. (1999)[78]. Some SMCs are action-focused ("bring-about SMCs") and others observation-focused ("classification SMCs"). A curriculum consisting of positive and negative examples of concepts are generated and used by human trainers in a 2D "PixelWorld" to facilitate concept learning in a hierarchical, layered manner. Exemplary concepts include containment, object, objects that are containers, pushability, being on top, and object number (i.e., being to the left of two objects). Pushability, for example, is built on six layers of lower level concepts. SMCs are invoked as functions that return a binary value indicating success or failure.

Other recent work from Vicarious includes: Lavin et al. (2018) on how reasoning about visual scenes learned using RCNs (George et al. 2017) can explain several well-known psychophysical and physiological results[79]; Lázaro-Gredilla et al. (2019) on *concepts as cognitive programs* that are learned from pairs of input-output images using a small set of primitive instructions (concepts, schemas) and run on a *visual cognitive computer* (VCC); Rikhye et al. (2020) on *clone-structured cognitive graphs* (CSCGs) for representing the environment and improving autonomous navigation by extending cloned hidden Markov models (HMMs) to include agent actions; Sawyer et al. (2020) on improving the cognitive program approach to concept learning (Lázaro-Gredilla et al. 2019) by adding human-inspired heuristics of object factorization and sub-goaling to accelerate program learning; Lázaro-Gredilla et al. (2021) on a *query training* (QT) method for learning *probabilistic graphical models* (PGMs) with paired inference algorithms to yield state-of-the-art results in applications like masked image region completion, learning full parameterization of grid-arranged Markov random fields, recognizing digits from noisy images, and other classification tasks.

[76] "Completely Automated Public Turing test to tell Computers and Humans Apart," commonly implemented using distorted or obfuscated characters to prevent non-human access to Web resources.
[77] This is technically accomplished by adding a loss term ("novel cost function") to the CNN that balances two feature maps: "one obtained from masking the input, and another derived from applying a mask in the feature space."
[78] In the Conclusion section of Vicarious (2018), it is suggested SMCs and Semantic Networks may be productively combined where schema networks "would allow the agent to have an internal representation of the external world that it can use for simulation and planning."
[79] These results included: subjective contours, neon color spreading, occlusion versus detection, and the border-ownership competition phenomenon. The reasoning method used was approximate Bayesian inference via loopy belief propagation.

## 5 Related Topics

Table 5.1 lists topics related to constructivist AI. The topics are grouped together and ordered roughly by decreasing relevance.

| Topics | References |
|---|---|
| Constructivist AI<br>AI constructivism<br>Developmental robotics<br>Autonomous mental development<br>Cognitive and developmental systems<br>Epigenetic robotics<br>Cognitive robotics<br>Computational approach to constructivism<br>Autonomous learning | See previous sections |
| Learning like a child<br>Learning like a baby<br>Building a baby<br>Learning like people | Turing 1950; Klahr and Wallace 1976; Ring 1994; Cohen PR et al. 1996; Guerin 2008a; Mao et al. 2015; Lake et al. 2016; Hutson 2018; Kwon 2018 |
| Continuous learning<br>Continual learning<br>Lifelong learning<br>Never-ending learning<br>Cumulative learning | Ring 1994[80]; Thrun and Mitchell 1995[81]; Ring et al. 2011; Chen Z and Liu 2016; Fei et al. 2016; DARPA 2017; Kirkpatrick et al. 2017; Mankowitz et al. 2018; Mitchell et al. 2018; Schwarz et al. 2018 |
| Cognitive architecture<br>Computational models<br>Hierarchical Temporal Memory<br>Biologically inspired cognitive architecture<br>Biological and machine intelligence | Newell 1973; Anderson and Kline 1977; Albus 1991, 1999; Schlesinger and McMurray 2012; Hawkins 2004, 2017; Samsonovich 2010; Laird et al. 2017; Kotseruba and Tsotsos 2020 |
| Concept learning<br>Concept formation<br>Surprise-based learning | Gennari et al. 1989; Fisher et al. 1991; Tenenbaum 1999; Barto et al. 2004; Ranasinghe and Shen 2008; Jia et al. 2013; Celikkanat et al. 2015b; MacLellan et al. 2016; Higgins I et al. 2017 |
| Knowledge bases<br>Ontologies | |
| Learning to learn<br>Meta-learning | Thrun and Pratt 1998; Schaul and Schmidhuber 2010; Lake 2016; Finn et al. 2017; Frans et al. 2017; Wang JX et al. 2017; Al-Shedivat et al. 2017; Rusu et al. 2018 |
| Reinforcement learning<br>Deep reinforcement learning<br>Model-based reinforcement learning<br>Intrinsically motivated learning | Sutton 1988; Barto et al. 2004; Oudeyer et al. 2007; Baldassarre and Mirolli 2013; Christiano et al. 2017; Arulkumaran et al. 2017; Weber et al. 2018; Sutton and Barto 2018 |
| Model building<br>Program induction | Lake et al. 2016 |
| Incremental learning<br>Online learning<br>Nonstationary learning | Gennari et al. 1989; Shen 1997; Utgoff and Stracuzzi 2002; Ditzler et al. 2015; Sutton and Barto 2018 |
| Commonsense knowledge acquisition | Guerin 2008b; Davis and Marcus 2015; Vicarious 2017c; Goyal et |

[80] Ring appears to have first referred to *continual learning/development* in his 1991 paper "Incremental Development of Complex Behaviors through Automatic Construction of Sensory-motor Hierarchies."
[81] Thrun and Mitchell describe an Explanation-Based Neural Network (EBNN) approach for enabling lifelong learning via a process consisting of Explain -> Analyze -> Learn steps. They suggest knowledge transfer from one domain to another is essential for lifelong learning with key transferrable elements being agent sensor and effector behaviors and any invariants learned about tasks and environments.

| | |
|---|---|
| | al. 2017 |
| Layered learning | Stone and Veloso 2000; Utgoff and Stracuzzi 2002; Dicarlo 2018; MacAlpine and Stone 2018 |
| Compositionality | Lake et al. 2016; Stone et al. 2017 |
| Bootstrap learning<br>Bootstrapping agent | Kuipers and Beeson 2002; Kuipers et al. 2006; Stober and Kuipers 2008; Dupoux 2016 |
| One-shot learning<br>Zero-shot learning<br>Few-shot learning<br>Zero-shot generalization | Cunningham 1972; Fei-Fei et al. 2006; Palatucci et al. 2009; Norouzi et al. 2014; Koch et al. 2015; Santoro et al. 2016; Ravi and Larochelle 2017; Kansky et al. 2017; Duan et al. 2017; Vinyals et al. 2017; Rusu et al. 2018 |
| Active learning | Cohn et al. 1996; Hasenjager and Ritter 2002; Bondu and Lemaire 2007; Settles 2010; Da Silva et al. 2014 |
| Transfer learning<br>Multi-task learning<br>Representation learning | Pan and Yang 2010; Lake et al. 2016; Oord et al. 2018 |
| Artificial General Intelligence<br>Seed AGI<br>Seed AI | Yudkowsky 2001; Adams et al. 2012 |
| Curriculum learning<br>Grounded-language learning | Elman 1993; Bengio et al. 2009; Hill et al. 2017 |
| Master Algorithm<br>Markov Logic Networks | Domingos 2015 |
| Learning disentangled factors | Higgins I et al. 2016 |
| Imitation learning<br>Learning by demonstration<br>Learning from observation | Schaal 1999; Friesen and Rao 2010; Chalodhorn and Rao 2010; Chung et al. 2014; Niekum et al. 2014; Nivel et al. 2014a; Duan et al. 2017 |
| Bottom-up learning | Sun and Zhang 2004 |
| Embodied cognition<br>Grounded cognition<br>Agent-based approach | Lakoff and Johnson 1980; Brooks et al. 1999; Schlesinger and Parisi 2001; Hedblom et al. 2015; Hedblom 2018 |
| Computational Creativity | Cohen LM 1989; Aguilar and Perez y Perez 2015, 2017 |
| Artificial Life<br>Computational Autopoiesis<br>Cellular Automata[82] | Gardner 1970; Wolfram 1984; McMullen and Varela 1997 |

Table 5.1. Topics related to constructivist AI.

Many other methods may be found in a constructivist AI toolkit. Shen (1994, section 3.4) suggests: function approximation, function optimization, classification and clustering, inductive inference and system identification, learning finite state machines and hidden Markov models, dynamic systems and chaos, problem solving and decision making, reinforcement learning, adaptive control, and developmental psychology. Appendix A lists dozens of other relevant topics from developmental psychology, neurobiology, and other disciplines.

## 6 Toward a General-Purpose Concept Learner

The author is pursuing AI constructivist development inspired by Piaget and informed by those surveyed in this paper. The goal is to engineer components capable of creating and maintaining memory structures (schemas) that enable agents to continually develop and refine *knowledge* and *skills* through interactions with their physical and social environments. Mechanisms supporting assimilation, accommodation, and staged development are being defined. The mechanisms are intended to enable cognitive development characteristic of lower animals through superhuman intelligences as a function of available resources and specified drives and goals. An eventual goal of this work is to configure an "infant bootstrap" agent that achieves human developmental milestones when provided with humanlike sensors and effectors and immersed in humanlike environments.

[82] These topics relate to discovering and implementing simple rules that combine to enable complex behaviors—consistent with constructivist AI schemas and staged development.

The initial focus is on generating and maintaining semantic and episodic memory structures that support concept and skill learning. Concepts are represented as *knowledge schemas* stored in semantic memory. Skills are represented as *sensorimotor schemas* stored in episodic memory. These are seen as necessary (but not sufficient) elements of a full-featured cognitive architecture.

Semantic memory will contain an ever-growing network of features/concepts[83] parsed from sensory and sensorimotor input streams using a General-Purpose Concept Learner (GPCL). Episodic memory will store episodes such as: (1) "Raw" sensory and sensorimotor episodes from which extracted features are derived (i.e., training cases), (2) Generalized scripts derived from similar (assimilated) episodes spanning a full range of agent skills (e.g., <grasping>, <subitizing>, <language processing>, <navigating>, <grocery shopping>, <logical reasoning>),[84] and (3) Episodes imagined by the agent (e.g., narratives). Episodes and extracted features will be linked to facilitate grounding, reasoning, retraining, and creative thought. Stored episodes will be available for replay so additional features can be extracted based on subsequent learning and so alternate focuses of attention can be applied.

Concept and skill development will occur concurrently, complimentarily, and continuously. Learning will be driven by intrinsic and extrinsic reinforcement and generally occur in an incremental and layered manner, resulting in a gradually accumulated hierarchy of concepts and skills. However, concepts and skills can emerge top-down as well as bottom-up. In the bottom-up case, they emerge from primitive features extracted from instances identified in episodes. For example, instances of seeing one's mother will induce a concept <my mother> (via assimilation) and, later, the categorical concept of <mothers> may emerge from examples of other instances and types of mothers (via accommodation). In the top-down case, more general concepts are introduced and serve as seeds to facilitate acquisition of lower-level concepts. Such higher-level concepts may be learned primarily through watching, imitating, and receiving external reinforcement from other agents. For example, in humans, concepts of <letters>, <words>, and <reading> are learned through interactions with parents before acquiring concepts of individual letters and words and the skill of reading. In non-human animals, concepts of <stalking>, <eluding>, and <hiding>, likely acquired by watching and interacting with other agents, precede the acquisition of specific techniques and skills related (subordinate) to <stalking>, <eluding>, and <hiding>. Such higher-level concepts enable and motivate agents to "fill in the gaps" in the wider conceptual areas.

Table 6.1 provides a simple example of knowledge schema labels at various levels of a knowledge hierarchy corresponding to six sensorimotor channels—from low- to high-level concepts. Each label represents a class consisting of a network of features parsed from and linked to episodes. Concepts and skills that persist (become reified) in semantic and episodic memory are those that prove most useful for subsequent reasoning and prediction. Other concepts and skills can be generated and forgotten on the fly in service of processing immediate sensorimotor inputs/outputs. Two key factors determining what features are learned are: (a) The biases (motivations) included in the learning algorithm and innate schemas (e.g., pain avoidance/pleasure attainment, survival, reproduction, discovering new concepts (curiosity), pleasing other agents, etc.), and (b) The order and nature of experienced episodes (i.e., curriculum learning).

[83] Features are used to induce concepts, which themselves are concepts induced from other features. "It's concepts all the way down."
[84] Thus episodic memory doubles as *procedural memory* suitable for encoding common procedures and plans. Elements in semantic and episodic memory are *reified* (i.e., persisted), in contrast to temporary versions that are processed in short-term memory during reasoning and other cognitive operations.

<table>
<tr><th>Vision</th><th>Audition</th><th>Touch</th><th>Olfaction</th><th>Taste</th><th>Intero-/ Proprio-ception</th></tr>
<tr><td>Edges<br>Segments<br>Textures<br>Luminosity<br>Brightness<br>Color<br>…</td><td>Phonemes<br>Timbre Pitch<br>Tempo<br>Rhythm<br>Volume</td><td>Painful<br>Hot Cold<br>Rough Gritty<br>Smooth<br>Hard Soft<br>Wet<br>Sharp</td><td>Key odor constituents</td><td>Sweet<br>Bitter<br>Sour<br>Salty<br>Umami</td><td>Balance<br>Nausea<br>Body-part positions</td></tr>
<tr><td><pleasant sight><br><unpleasant sight><br><big thing><br><liquid sight></td><td><pleasant sound> <unpleasant sound> <loud> <growl> <yell> <clap> <explosion></td><td><pleasant touch><br><unpleasant touch><br><solid touch><br><liquid touch></td><td><pleasant scent><br><unpleasant scent></td><td><pleasant taste><br><unpleasant taste></td><td><dizzy></td></tr>
<tr><td><see arm move><br><mom sight><br><joes guitar><br><guitar><br><persistent object><br><A> <B> <C><br><word> <sentence><br><reading></td><td><mom sound> <human sound> <cow sound><br><english speech><br><foreign speech><br><singing><br><music><br><jazz> <pop></td><td><feel arm move><br><mom feel></td><td><mom scent><br><rose scent></td><td><chocolate><br><milk choc><br><dark choc><br><black pepper><br><hot sauce></td><td></td></tr>
<tr><td colspan="6">Persistent objects <move_arm> <grasp> <close> <far away> <nice_thing> <caregiver> <my mother> <containment> <pathway> <my dog> <joes dog> <crawling> <powerful being> <happy> <sad> <I did good> <I did bad> <funny> <left> <right> <forward> <back> <up> <down> <subitizing> <estimating></td></tr>
<tr><td colspan="6"><smart phone> <using smart phone> <somersault> <reading> <back flip> <skiing> <counting></td></tr>
<tr><td colspan="6"><comparing size> <addition> <subtraction></td></tr>
<tr><td colspan="6"><mothers> <motherhood> <dogs> <parent> <god> <arithmetic> <sharing> <ownership> <cooperation> <competition> <empathy> <multiplying> <dividing> <measuring with stick></td></tr>
<tr><td colspan="6"><playing chess> <glenn gould> <bachs art of the fugue> <art of the fugue performed by glenn gould></td></tr>
<tr><td colspan="6"><mathematics> <logic> <democracy> <freedom></td></tr>
<tr><td colspan="6"><physics> <chemistry> <biology> <liberal arts> <natural selection></td></tr>
<tr><td colspan="6"><supervised learning> <unsupervised learning> <reinforcement learning> <set theory></td></tr>
<tr><td colspan="6"><ai constructivism> <philosophy> <existentialism> <quantum mechanics> <big history> <cosmology></td></tr>
</table>

Table 6.1. Possible concepts in a knowledge hierarchy.

Concepts exist separate from language. Agents that intrinsically have or can learn language skills get a great cognitive boost from the ability to associate symbols with concepts. Words allow concepts to be introduced by "teachers" (i.e., assigned class labels) and skills acquired without the agent needing to be explicitly shown or experience the skill. Once associated with a concept, words serve as just another predictive feature linked to underlying schema concepts. Features that are learned subconsciously may later be accessed consciously, symbolically labeled (named), and used in creating new schemas. For example, <green patch> or <left loop> features discovered in doing visual object recognition could be named later and leveraged for additional concept formation.

As another example, an agent may learn the concept <wooly texture> (unconsciously) as a byproduct of learning the concepts <sheep>, <alpaca>, and <labradoodle> before learning the word "wooly" and (consciously) associating it with <wooly texture>. A dog may master the action <chew up master's fuzzy slippers> before learning <bad dog> and "bad dog" (simultaneously). When the dog later executes the action <defecate in master's sleeping area>, the <bad dog> concept quickly assimilates the new concept into the <bad dog> concept upon hearing and experiencing the master's "bad dog" reaction. An example of word learning occurring before concept learning may be illustrated by the word "quantum mechanics" being introduced to an agent before the agent has knowledge of the concept <quantum mechanics>. As concepts associated with <quantum mechanics> are taught, the agent builds the concept of <quantum

mechanics> into a network of other concepts. One might think of semantic memory as constituted by Large Concept Models with one or more Large Language Models overlaid/integrated therein.

The GPCL framework presumes motor control is not strictly necessary for concept development. Much can be learned from purely sensory inputs *given suitable innate concepts and (mental) skills*[85]--say, an innate skill for doing object detection for example. One imagines an Intelligent Traffic Camera that can learn concepts about vehicles, environmental conditions, and other objects and actions just through appropriate parsing of its video input—without needing physical panning and zoom capabilities. This allows knowledge schemas to be developed without implementing (or simulating) motor controls. Companion sensorimotor schemas can be used and developed without needing the integrated motor aspects (a.k.a. *sensory schemas* or *cognitive skills*). That is, purely sensory streams (movies) consisting of one or more of video, audio, touch, taste, and olfaction, can be effectively parsed to yield significant levels of cognition. GPCL focuses on learning knowledge schemas and does not currently intend to implement sensorimotor learning[86].

Although it draws on developmental psychology and cognitive neuroscience, the GPCL approach is AI-oriented and leverages basic machine learning principles. Specifically, it plans to implement a hybrid of supervised and unsupervised learning (classification and clustering) driven by intrinsic and extrinsic reinforcement. It does not set out to model actual biological mechanisms. The memory structures and learning algorithms must support a practically unlimited number of episodes and concepts—as is the case in biological brains.

## Appendix A: Cangelosi and Schlesinger (2015) Topics, Keywords, and References

The following table summarizes topics and references covered in Cangelosi and Schlesinger's (2015) book *Developmental Robotics: From Babies to Robots*. It reflects work in developmental psychology and developmental robotics that was (mostly) not covered in this survey.

| **Topics and *keywords*** | **References** |
|---|---|
| Novelty, curiosity, and surprise (Ch.3) *intrinsic motivation (IM), knowledge-based IM, competence-based IM, prediction-based IM, extrinsic motivation, internal motivation, effectance, self-efficacy, contingency perception, personal causation, self-determination theory, functional assimilation, dopamine release in mesolimbic pathway, frontal eye field activity, superior colliculus activation, exploratory behavior, novelty detection, moderate novelty principle, exogenous orienting, habituation-dishabituation, comparator theory, internal templates, visual expectation paradigm (VExP), anticipations, object permanence, spontane-* | Hull 1943; Harlow 1950; Butler 1953; Kish and Anonitis 1956; White RW 1959; Berlyne 1960; Sokolov 1963; Hunt 1965; de Charms 1968; Smilansky 1968; Hunt 1970; Kagan 1972; Wetherford and Cohen 1973; Vinogradova 1975; Fischer KW 1980; Rovee-Collier and Sullivan 1980; Bahrick and Watson 1985; Deci and Ryan 1985; Bandura 1986; Ginsburg and Opper 1988; Haith et al. 1988; Johnson MH 1990; Bronson 1991; Canfield and Haith 1991; Schmidhuber 1991b; Haith et al. 1993; Gergely and Watson 1999; Dannemiller 2000; Roder et al. 2000; Ryan and Deci 2000; Horvitz 2000; Berthier et al. 2001; Bjorklund and Pellegrini 2002; Huang and Weng 2002; Gilmore and Thomas 2002; Wentworth et al. 2002; Johnson SP et al. 2003a; Barborica and Ferrera 2004; Barto et al. 2004; Marshall J et al. 2004; Sirois and Mareschal 2004; Colombo and Cheatham 2006; Redgrave and Gurney 2006; Kumaran and Maguire 2007; Matsumoto et al. 2007; Oudeyer et al. 2007; Schembri et al. 2007; Vieira-Neto and Nehmzow 2007; Adler at al. 2008; Fiore et al. 2008; Hiolle and Cañamero 2008; Isoda and Hikosaka 2008; Ryan and Deci 2008; Bromberg-Martin and Hikosaka 2009; Colombo and Mitchell 2009; Lee et al. 2009; Ferrera and Barborica 2010; Merrick 2010; Baldassarre 2011; Wright and Panksepp 2012; Gottlieb et al. 2013; Mirolli and Baldassarre 2013 |

[85] Innate concepts and skills may be provided by preconfigured knowledge and sensorimotor schemas. They are expected to provide the fundamental goals and drives (foundational concepts) that direct subsequent development. In biological agents, such concepts and skills are learned through evolution and activated postnatally by appropriate sensations.

[86] This is not to suggest sensorimotor learning is less important. It simply reflects the development focus of this project. The work of others will be drawn on for sensorimotor schema learning such as circular operations, backward and forward models, and many other approaches noted in this paper. However, classes of powerful cognitive agents can likely be created that have minimal motor abilities and learn exclusively through sensory inputs.

| | |
|---|---|
| *ous play, reinforcement learning (RL), uncertainty motivation, information-gain motivation, predictive-novelty motivation, maximizing-incompetence motivation, learning-progress motivation; maximizing-competence progress, salience maps, options framework, actor-critic architecture* | |
| <u>Visual development (Ch.4)</u> *experience-expectant development, moderate novelty principle, architectural innateness, preferential looking, habituation-dishabituation, intermodal contingencies, face recognition, spatial perception, optic flow, egocentric-allocentric bias, self-perception, mirror self-recognition, object recognition (texture, shape, common surfaces, perceptual completion, unity perception), learning affordances, tool use, forward-inverse model* | James 1890; Piaget 1952; Fantz 1956; Gibson and Walk 1960; Charlesworth 1969; Campos et al. 1970; Gallup 1970; Maurer and Salapatek 1976; Meltzoff and Moore 1977; Acredolo 1978; Lewis and Brooks-Gunn 1979; Meltzoff and Borton 1979; Acredolo and Evans 1980; Haith 1980; Maurer and Barrera 1981; Marr 1982; Acredolo et al. 1984; Bahrick and Watson 1985; Gibson 1986; Kestenbaum et al. 1987; Kermoian and Campos 1988; Bushnell IWR et al. 1989; Spelke 1990; Ballard 1991; Morton and Johnson 1991; Campos et al. 1992; Butterworth 1992; Bushnell EW and Boudreau 1993; Leinbach and Fagot 1993; van Leeuwen et al. 1994; Bahrick et al. 1996; Elman et al. 1996; Higgins C et al. 1996; Slater et al. 1996; Valenza et al. 1996; Adolph 1997; Hiraki et al. 1998; Rochat 1998; Asada et al. 1999; Greenough and Black 1999; Schlesinger and Langer 1999; Itti and Koch 2000; Bushnell IWR 2001; McCarty et al. 2001b; Weng et al. 2001; Meissner and Brigham 2001; Bednar and Miikkulainen 2002; Courage and Howe 2002; de Haan et al. 2002; Furl et al. 2002; Mareschal and Johnson 2002; Quinn et al. 2002; Rochat and Striano 2002; Bednar and Miikkulainen 2003; Cos-Aguilera et al. 2003; Johnson SP et al. 2003a; Johnson SP et al. 2003b; Chen Y and Weng 2004; Fitzpatrick and Arsenio 2004; Johnson SP 2004; Lovett and Scassellati 2004; Michel et al. 2004; Natale et al. 2005b; Stoytchev 2005; Amso and Johnson 2006; Fritz et al. 2006; Joh and Adolph 2006; Zhang and Lee 2006; Fuke et al. 2007; Sahin et al. 2007; Sann and Streri 2007; Schlesinger et al. 2007; Schlesinger and Parisi 2007; Ugur et al. 2007; Adolph 2008; Fitzpatrick et al. 2008; Guerin and McKenzie 2008; Montesano et al. 2008; Stoytchev 2008; Sturm et al. 2008; Gold and Scassellati 2009; Kaipa et al. 2010; Franz and Triesch 2010; Chaudhuri 2011; Stoytchev 2011; Schlesinger et al. 2012 |
| <u>Motor development (Ch. 5)</u> *learning motor skills, u-shaped development, motor babbling, body babbling, multiple sense integration, experience-expectant development, experience-dependent development, pre-reaching, pre-shaping, grasp development (palmar, scissors, radial-digital, pincer, mature pincer), locomotion development (pre-crawling, creeping, crawling, sideways cruising, frontward cruising, standing, walking), central pattern generator (CPG), CPG models, exploration-tuning-mastery, stiffness to fluidity, INFANT model, freezing degrees of freedom, speed-accuracy tradeoff, visuomotor mapping* | Gesell 1945; Gesell 1946; McGraw 1945; White BL et al. 1964; Bower et al. 1970; Zelazo et al. 1972; Trevarthen 1975; Field 1977; McDonnell and Abraham 1979; von Hofsten 1982; Lockman et al. 1984; von Hofsten 1984; von Hofsten and Fazel-Zandy 1984; Bushnell EW 1985; Thelen 1986; Thelen et al. 1987; Kermoian and Campos 1988; Kuperstein 1988; Goldfield 1989; Newell et al. 1989; Kuperstein 1991; Thelen and Ulrich 1991; Bril and Breniere 1992; Ashmead et al. 1993; Bullock et al. 1993; Clark JE and Phillips 1993; Clifton et al. 1993; Goldfield et al. 1993; Sporns and Edelman 1993; von Hofsten and Rönnqvist 1993; Vos and Scheepstra 1993; Erhardt 1994; Freedland and Bertenthal 1994; Berthier 1996; Ennouri and Bloch 1996; Johnson and Blasco 1997; Konczak and Dichgans 1997; Adolph et al. 1998; Hase and Yamazaki 1998; McCarty and Ashmead 1999; McCarty et al. 1999; Metta et al. 1999; Smeets and Brenner 1999; Vereijken and Adolph 1999; Arena 2000; Schlesinger et al. 2000; McCarty et al. 2001a; McCarty et al. 2001b; Weng et al. 2001; Wheeler et al. 2002; Lungarella and Berthouze 2003; Berthouze and Lungarella 2004; Oztop et al. 2004; Shadmehr and Wise 2004; Berthier and Keen 2005; Berthier et al. 2005; Natale et al. 2005a; Witherington 2005; Kuniyoshi and Sangawa 2006; Righetti and Ijspeert 2006a, Righetti and Ijspeert 2006b; Taga 2006; Degallier et al. 2007; Lee et al. 2007; Natale et al. 2007; Nori et al. 2007; von Hofsten 2007; Caligiore et al. 2008; Degallier et al. 2008; Ijspeert 2008; Asada et al. 2009; Wu QD et al. 2009; Gerber et al. 2010; Hulse et al. 2010a; Hulse et al. 2010b; Metta et al. 2010; Berthier 2011; Law et al. 2011; Li et al. 2011; Lu et al. 2012; Savastano and Nolfi 2012; Li et al. 2013 |
| <u>Social learning (Ch. 6)</u> *eye contact, joint attention, gestures (deictic, representational), gaze following (sensitivity, ecological, geometrical, and representational stages), pointing (imperative, declarative), im-* | Field et al. 1983; Meltzoff and Moore 1983; Meltzoff 1988; Meltzoff and Moore 1989; Butterworth 1991; Butterworth and Jarrett 1991; Leslie 1994; Baron-Cohen 1995; Davies and Stone 1995; Fadiga et al. 1995; Meltzoff 1995; Demiris et al. 1997; Meltzoff and Moore 1997; Rizzolatti and Arbib 1998; Wolpert and Kawato 1998; Iacoboni et al. 1999; Nadel and Butterworth 1999; Scassellati 1999; Schaal 1999; Billard and Matarić 2001; Ishiguro et al. 2001; Kozima and Yano 2001; Rizzolatti et al. 2001; Breazeal and Scassellati 2002; |
| *itation, collaboration/ cooperation, intention reading, goal prediction, theory of mind, self-recognition, AIM model, mirror neurons, MOSAIC model, HAMMER architecture, paired inverse-forward models, world model* | Call and Carpenter 2002; Demiris and Hayes 2002; Fasel et al. 2002; Nehaniv and Dautenhahn 2002; Scassellati 2002; Demiris and Johnson 2003; Gergely 2003; Imai et al. 2003; Nagai et al. 2003b; Nagai et al. 2003b; Ude and Atkeson 2003; Carlson E and Triesch 2004; Ito and Tani 2004; Rizzolatti and Craighero 2004; Zöllner et al. 2004; Borenstein and Ruppin 2005; Breazeal et al. 2005; Carpenter et al. 2005; Demiris and Deardon 2005; Hafner and Kaplan 2005; Thomaz et al. 2005; Tomasello et al. 2005; Demiris and Khadhouri 2006; Demiris and Simmons 2006; Dominey et al. 2006; Ferrari et al. 2006; Hashimoto et al. 2006; Kaplan and Hafner 2006; Mavridis and Roy 2006; Nagai et al. 2006; Triesch et al. 2006; Warneken et al. 2006; Warneken and Tomasello 2006; Meltzoff 2007; Nehaniv and Dautenhahn 2007; Rao et al. 2007; Tapus et al. 2007; Watanabe et al. 2007; Barsalou 2008; Call and Tomasello 2008; Demiris and Meltzoff 2008; Jasso et al. 2008; François et al. 2009a; François et al. 2009b; Gold and Scassellati 2009; Tomasello 2009; Kruger et al. 2010; Lallée et al. 2010; Dominey and Warneken 2011; Hafner and Schillaci 2011; Sarabia et al. 2011; Tanz 2011; Carlson T and Demiris 2012 |
| Language development (Ch. 7)<br>*nature vs nurture, nativism vs empiricism, language acquisition device, grammaticalization, usage-based theory of language development, cognitive linguistic theories, symbol grounding, babbling, vocabulary spurt, verb islands, biases (reference, similarity, conventionality, whole-object, whole-part juxtaposition, segmentation, taxonomic, mutual exclusivity, embodiment, social cognition), Modi experiment, linguistically enabled synthetic agent (LESA), phoneme-syllabic-lexical layer models, epigenetic robotics architecture, fluid construction grammar (FCG), semantic compositionality* | Chomsky 1957; Chomsky 1965; Bloom L 1973; Braine 1976; McClelland and Rumelhart 1981; Markman and Hutchinson 1984; Brooks 1986; Flege 1987; Langacker 1987; Mervis 1987; Markman and Wachtel 1988; Gleitman 1990; Harnad 1990; Plunkett et al. 1992; Tomasello 1992; Baldwin D 1993; Clark EV 1993; Fenson et al. 1994; Golinkoff et al. 1994; Pinker 1994; Berrah et al. 1996; Elman et al. 1996; Regier 1996; Saffran et al. 1996; Vihman 1996; Carpenter et al. 1998; MacWhinney 1998; Barrett 1999; Jusczyk 1999; Tomasello and Brooks 1999; Bloom P 2000; Browman and Goldstein 2000; Cangelosi et al. 2000; Oller 2000; Christiansen and Chater 2001; de Boer 2001; Kirby 2001; Cangelosi and Parisi 2002; Saylor et al. 2002; Steels and Kaplan 2002; Pulvermüller 2003; Steels 2003; Tani 2003; Tomasello 2003; Yoshikawa et al. 2003; Roy et al. 2004; Bortfeld et al. 2005; Dominey and Boucher 2005a; Dominey and Boucher 2005b; Iverson and Goldin-Meadow 2005; Pecher and Zwaan 2005; Samuelson and Smith 2005; Smith 2005; Sugita and Tani 2005; Yu 2005; Cangelosi and Riga 2006; Ho et al. 2006; Goldberg 2006; Ogino et al. 2006; Oudeyer 2006; Oudeyer and Kaplan 2006; Steels and de Beule 2006; Hornstein and Santos-Victor 2007; Lopes and Chauhan 2007; Mareschal et al. 2007; McMurray 2007; Tomasello et al. 2007; Baldwin and Meyer 2008; Brandl et al. 2008; CMU 2008; Hofe and Moore 2008; Mikhailova et al. 2008; ten Bosch and Boves 2008; Tomasello 2008; Carpenter 2009; Driesen et al. 2009; Hoff 2009; Ishihara et al. 2009; Cangelosi 2010; Cangelosi et al. 2010; Gläser and Joublin 2010; Lyon et al. 2010; Marocco et al. 2010; Mayor and Plunkett 2010; Morse et al. 2010a; Morse et al. 2010b; Smith and Samuelson 2010; Tuci et al. 2010; Laurent et al. 2011; Rothwell et al. 2011; Tikhanoff et al. 2011; Steels 2011; Lyon et al. 2012; Mangin and Oudeyer 2012; Spranger 2012a; Spranger 2012b; Steels 2012; Tallerman and Gibson 2012; Araki et al. 2013; Peelle et al. 2013 |
| Abstract knowledge (Ch. 8)<br>*Cognitive computational models, embodied robotic models, number cognition, set discrimination, cardinality principles, order/item irrelevance principles, number directionality, embodied number cognition effects (size and distance, SNARC, Posner-SNARC, context and function, subitizing (1-3), gesture and finger counting), entity concept, concrete-to-abstract pathway, abstract words and symbols, mode of acquisition (MoA), function words, negation concepts, linguistic generativity, symbol grounding transfer, representations for decision-making, internal simulations, ISAC architecture, iCub architecture, other cognitivist/ emergentist/ hybrid* | Kaufman et al. 1949; Piaget 1952; Piaget 1972; Spitz 1957; Bloom L 1970; Austin 1975; Gelman and Tucker 1975; Pea 1978; Pea 1980; Starkey and Cooper 1980; von Hofsten 1982; Gelman et al. 1986; Choi 1988; Keil 1989; Gelman 1990; Harnad 1990; Wynn 1990; Schwanenflugel 1991; Wynn 1992; Bates and Elman 1993; Dehaene et al. 1993; Moxey and Sanford 1993; Simons and Keil 1995; Starkey and Cooper 1995; Cowan et al. 1996; Dehaene 1997; Schwarz and Stein 1998; Alibali and DiRusso 1999; Barsalou 1999; Graham 1999; Rodriguez et al. 1999; Lakoff and Núñez 2000; Xu F and Spelke 2000; Wiemer-Hastings et al. 2001; Sun et al. 2001; Parisi and Schlesinger 2002; Wakeley et al. 2000; Wang H et al. 2001; Fischer MH et al. 2003; Wauters et al. 2003; Caramelli et al. 2004; Barsalou and Wiemer-Hastings 2005; Bisanz et al. 2005; Campbell 2005; Cordes and Gelman 2005; Coventry et al. 2005; Rajapakse et al. 2005; Verguts et al. 2005; Cangelosi and Riga 2006; Andres et al. 2007; Cangelosi et al. 2007; Sun 2007; Tikhanoff et al. 2007; Vernon et al. 2007; Barsalou 2008; Fischer MH 2008; Kawamura et al. 2008; Thornton 2008; Langley et al. 2009; Borghi and Cimatti 2010; Caligiore et al. 2010; Chen Q and Verguts 2010; Della Rosa et al. 2010; Gordon et al. 2010; Vernon et al. 2010; Borghi et al. 2011; Förster et al. 2011; Rucinski, Cangelosi, and Belpaeme 2011; Kostas et al. 2011; Saunders et al. 2011; Rucinski et al. 2012; Lyon et al. 2012; Stramandinoli et al. 2012; Förster 2013 |
| *architectures (25+ mentioned)* | |